%% file: main.tex
\documentclass{article}
\usepackage[utf8]{inputenc} %
\usepackage[T1]{fontenc}    %
\usepackage{mdframed}
\usepackage{verbatim}
\usepackage{graphicx}
\usepackage{subcaption}
\usepackage{booktabs} %
\usepackage{multirow}
\usepackage{tabularx}
\usepackage[table]{xcolor}
\usepackage{wrapfig}

\usepackage{hyperref}
\hypersetup{hidelinks}

\PassOptionsToPackage{numbers}{natbib}
\usepackage[final,eandd]{neurips_2026}
\setcitestyle{authoryear,round,citesep={;},aysep={,},yysep={;}}
\usepackage{amsmath}
\usepackage{amssymb}
\usepackage{mathtools}
\usepackage{amsthm}

\usepackage{needspace}
\usepackage{placeins}

\usepackage[capitalize,noabbrev]{cleveref}

\theoremstyle{plain}

\theoremstyle{definition}

\theoremstyle{remark}

\usepackage[textsize=tiny]{todonotes}
\usepackage{paralist}

\usepackage{xstring}
\newcommand{\score}[1]{%
  \StrDel{#1}{.}[\cleanscore]%
  \ifnum\cleanscore>8750\cellcolor{green!30}#1\else
  \ifnum\cleanscore>7000\cellcolor{green!20}#1\else
  \ifnum\cleanscore>5000\cellcolor{yellow!20}#1\else
  \ifnum\cleanscore>3750\cellcolor{yellow!30}#1\else
  \ifnum\cleanscore>1250\cellcolor{orange!30}#1\else
  \cellcolor{red!20}#1\fi\fi\fi\fi\fi
}

\newcommand{\bestscore}[1]{%
  \StrDel{#1}{.}[\cleanscore]%
  \ifnum\cleanscore>8750\cellcolor{green!30}\textbf{#1}\else
  \ifnum\cleanscore>7000\cellcolor{green!20}\textbf{#1}\else
  \ifnum\cleanscore>5000\cellcolor{yellow!20}\textbf{#1}\else
  \ifnum\cleanscore>3750\cellcolor{yellow!30}\textbf{#1}\else
  \ifnum\cleanscore>1250\cellcolor{orange!30}\textbf{#1}\else
  \cellcolor{red!20}\textbf{#1}\fi\fi\fi\fi\fi
}

\newcommand{\topic}[1]{\cellcolor{blue!20}\textbf{#1}}
\newcommand{\topicSystem}[1]{\cellcolor{blue!10}\textbf{#1}}
\newcommand{\topicAI}[1]{\cellcolor{green!10}\textbf{#1}}
\newcommand{\topicApplication}[1]{\cellcolor{yellow!10}\textbf{#1}}
\newcommand{\topicOthers}[1]{\cellcolor{red!10}\textbf{#1}}

\definecolor{waterblue}{RGB}{173, 216, 230}

\newcommand{\diffpos}[1]{{\scriptsize\color{green!60!black}(+#1\%)}}
\newcommand{\diffneg}[1]{{\scriptsize\color{red!60!black}(-#1\%)}}
\newcommand{\diffzero}{{\scriptsize\color{gray}(+0.00\%)}}
\definecolor{colorStrat}{RGB}{40, 110, 230} 
\definecolor{colorDebug}{RGB}{210, 80, 20}
\definecolor{colorEnv}{RGB}{180, 140, 0}
\definecolor{colorImpl}{RGB}{20, 140, 130}

\newcommand{\failStrat}[1]{{\color{colorStrat}#1}}
\newcommand{\failDebug}[1]{{\color{colorDebug}#1}}
\newcommand{\failEnv}[1]{{\color{colorEnv}#1}}
\newcommand{\failImpl}[1]{{\color{colorImpl}#1}}
\begin{document}
\raggedbottom

\title{CSBench: Evaluating Project-Level System Construction in Computer Science}
\author{%
  \normalfont 
  Hongli Yu\textsuperscript{1,2,3}\thanks{\protect\hypertarget{Hfootnote.1}{}Equal contribution.}\quad Huang-ang Gao\textsuperscript{1,2,3\hyperlink{Hfootnote.1}{*}}\quad Botian Wang\textsuperscript{1,3\hyperlink{Hfootnote.1}{*}}
  \quad Hanlin Wu\textsuperscript{1,2,3} \\
  YiFei Wang\textsuperscript{1,3}\quad Daoguang Zan\textsuperscript{2}\quad Anni Huang\textsuperscript{2}\quad Weinan Dai\textsuperscript{1,2,3} \\
  Qiying Yu\textsuperscript{1,2,3}\quad Jiaze Chen\textsuperscript{2,3}\quad Wei-Ying Ma\textsuperscript{1,3}\quad Ya-Qin Zhang\textsuperscript{1,3} \\
  Jingjing Liu\textsuperscript{1,3}\quad 
  Mingxuan Wang\textsuperscript{2,3}\thanks{\protect\hypertarget{Hfootnote.2}{}Corresponding authors.}\quad Hao Zhou\textsuperscript{1,3\hyperlink{Hfootnote.2}{\textdagger}} \\[0.5em]
  \normalfont\small \textsuperscript{1}Institute for AI Industry Research (AIR), Tsinghua University \\
  \normalfont\small \textsuperscript{2}ByteDance Seed \\
  \normalfont\small \textsuperscript{3}SIA-Lab of Tsinghua AIR and ByteDance Seed
}
\begingroup
\hypersetup{hidelinks}
\maketitle
\endgroup

\begin{abstract}
While LLM-based coding agents have advanced to repository-scale engineering, existing benchmarks predominantly focus on software maintenance, overlooking the foundational capability of \textit{system construction}. We introduce \textbf{CSBench}, a benchmark evaluating project-level Spec-to-Code implementation across core computer science domains (e.g., OS, compilers, databases). CSBench features 100 expert-curated tasks from top-tier university assignments within containerized environments. Evaluating 9 state-of-the-art LLMs reveals a stark performance disparity: even the strongest model, GPT-5.2, achieves 94.44\% on Machine Learning tasks but only 34.78\% on the broad System category. Our analysis highlights persistent challenges in \textbf{systemic reasoning} and \textbf{construction knowledge}. Agents often stop before completing all requirements or fail to use execution feedback to identify and fix underlying problems. These findings motivate CSBench as a testbed for evaluating and improving system construction capabilities, advancing autonomous agents toward deep computer science mastery.
\end{abstract}
\begin{figure*}[h]
  \centering
  \includegraphics[width=0.88\linewidth]{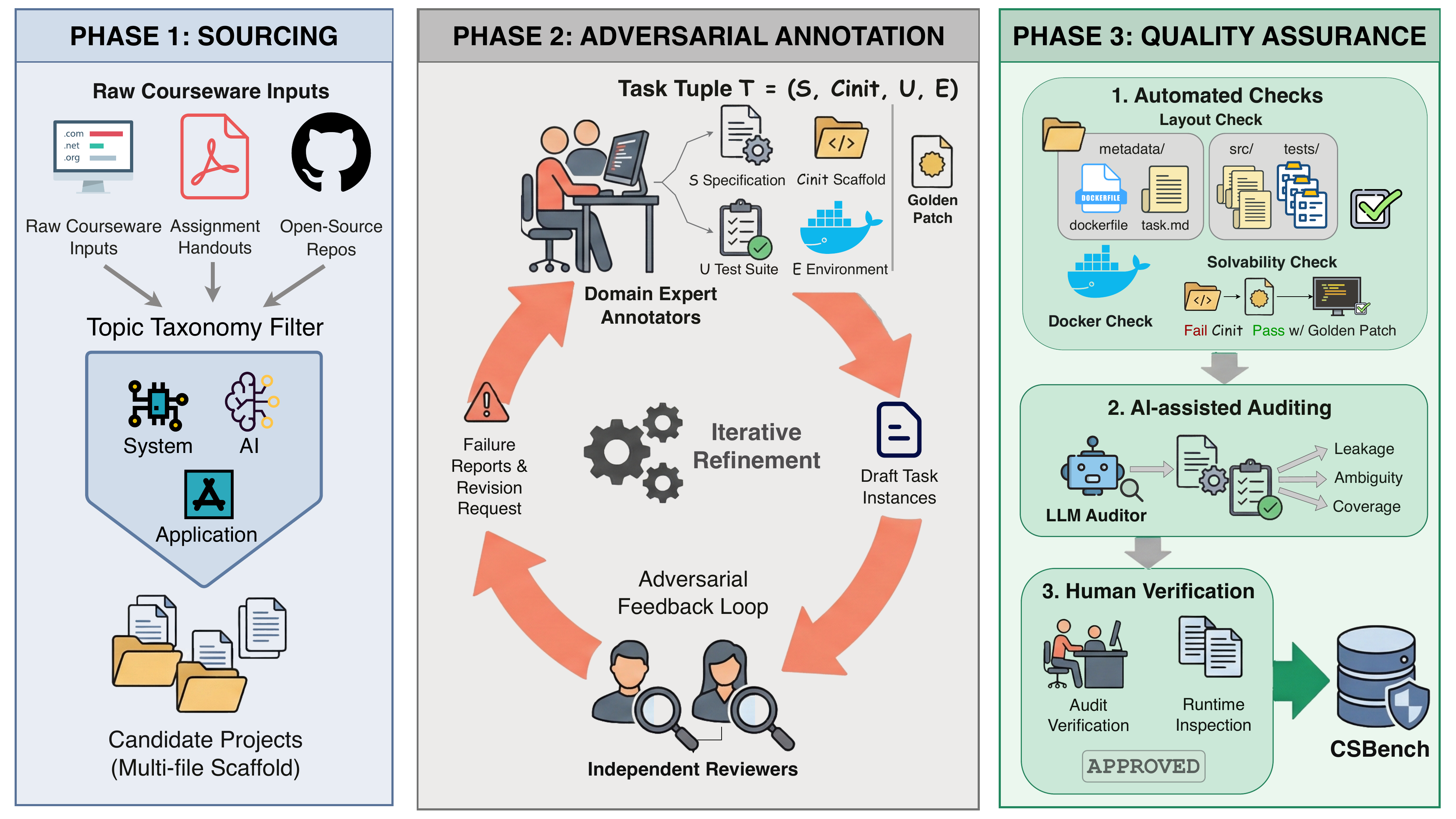}
  \caption{\textbf{Overview of CSBench.} CSBench evaluates project-level Spec-to-Code system construction from rigorous specifications. It is built through a three-stage pipeline of broad-scope project sourcing, adversarial expert annotation, and multi-layer quality assurance.}
  \label{fig:construction_pipeline}
\end{figure*}

\newcolumntype{Y}{>{\centering\arraybackslash}X}

\section{Introduction}
\label{sec:intro}

The evolution of Large Language Models (LLMs) for code has rapidly transitioned from generating isolated snippets to engineering at the repository scale. Early benchmarks like HumanEval \citep{Chen2021EvaluatingLL} and MBPP \citep{Austin2021ProgramSW} focused on algorithmic puzzles within single functions. However, real-world software engineering demands far more: understanding complex dependencies, navigating large codebases, and implementing features that span multiple files. This shift has driven the development of repository-level agents capable of autonomous planning and execution.

To evaluate these capabilities, the community has introduced benchmarks grounded in real-world repositories. SWE-bench \citep{Jimenez2023SWEbenchCL} pioneered this direction by assessing an agent's ability to resolve GitHub issues, primarily focusing on \textit{software maintenance}. More recently, SWE-Dev \citep{wang2025swe} extended this scope to \textit{Feature-Driven Development}, challenging agents to add features to existing projects. Despite these advances, these benchmarks predominantly test an agent's ability to \textit{adapt} to existing code-bases, such as following established patterns to fix bugs or add incremental features. They often overlook a more fundamental capability: \textbf{System Construction}, where agents must interpret detailed requirements, reason about domain-specific algorithms and abstractions, and assemble functional systems rather than merely modify existing ones.

To address this gap, we introduce \textbf{CSBench}, a benchmark for evaluating \textit{Spec-to-Code} construction of core computer science systems. Unlike existing repository-level benchmarks \citep{Jimenez2023SWEbenchCL, wang2025swe, zan2025multi} that primarily focus on adapting established codebases, CSBench challenges agents to implement compilers, database engines, operating system kernels, and other systems from rigorous specifications.

CSBench is characterized by three key features.
(i) \textbf{Broad-Scope Domain Coverage}: CSBench contains a 100-task evaluation set spanning 19 core computer science topics, including Operating Systems, Databases, Compilers, and AI, enabling evaluation of broad CS mastery beyond language proficiency.
(ii) \textbf{Construction-Oriented Reasoning}: Tasks require agents to implement complex algorithms and data structures, such as B+ Trees and TCP stacks, from scaffolded codebases, demanding long-horizon planning, requirement tracking, and implementation consistency.
(iii) \textbf{Pedagogical Rigor}: Sourced from top-tier university assignments, each task includes an expert-curated test suite designed to verify substantive functional correctness and catch corner cases.

We evaluate 9 frontier LLMs on CSBench and observe substantial performance differences across computer science domains. The top-performing model, GPT-5.2, achieves 94.44\% on Machine Learning but 34.78\% on the broad System category; among system topics with at least five tasks, it obtains 37.50\% on Operating Systems, 20.00\% on Parallel \& Distributed Systems, and 19.44\% on System Fundamentals. The failure analysis further shows that requirement-tracking and feedback-driven debugging failures are common in the evaluated trajectories. We find that agents struggle to maintain logical consistency over the long implementation trajectories required to build functional systems. They often stop before completing all requirements or fail to use execution feedback to identify and fix underlying problems. CSBench evaluates these capabilities by testing whether agents can translate detailed specifications into system construction. We open-source the CSBench tasks and evaluation code to support reproducible research.\footnote{\urlstyle{same}Code: \url{https://github.com/BytedTsinghua-SIA/CSBench}. Project website: \url{https://csbench-sia.github.io/}.}

Our contributions are summarized as follows:
\begin{itemize}
    \item We introduce CSBench, a benchmark focusing on Spec-to-Code system construction, filling the gap left by maintenance-centric evaluations.
    \item We curate 100 diverse tasks with canonical solutions and containerized evaluation environments, enabling reproducible evaluation of project-level system construction.
    \item We provide a comprehensive analysis of current SOTA models, offering new insights into their capabilities and limitations in applying systematic computer science knowledge.
\end{itemize}

\begin{figure*}[t]
    \centering
    \includegraphics[width=0.92\linewidth]{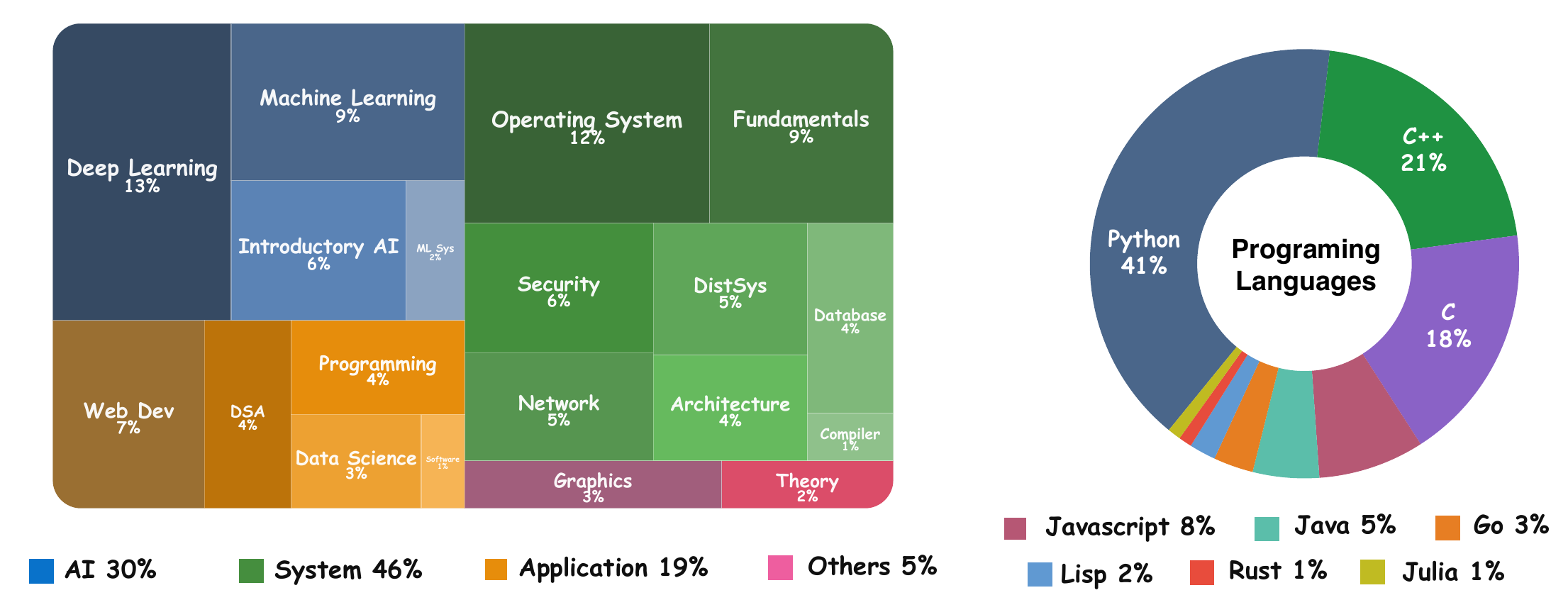}
    \caption{Left: task distribution of the 100-task evaluation set by course topic taxonomy (Systems/AI/Applications/Others) with the name and percentage of subtopics displayed. Right: programming language distribution across CSBench tasks.}
    \label{fig:category&language}
\end{figure*}

\section{CSBench}
In this section, we introduce the design and curation of CSBench, a rigorous benchmark for evaluating the \textbf{agentic system construction} capabilities of LLMs. Unlike existing benchmarks~\citep{Jimenez2023SWEbenchCL,wang2025swe,zan2025multi} that primarily focus on software maintenance (e.g., bug fixing within established repositories), CSBench targets \textbf{Spec-to-Code System Construction}. By leveraging expert-curated computer science projects, our benchmark simulates a scaffolded development environment where LLM agents must implement functional, system-level components from formal specifications, requiring deep domain knowledge and long-horizon planning.

The remainder of this section is organized as follows: \cref{sec:task_definition} provides a formal definition of the CSBench task instance and its evaluation protocol. We then detail the data curation and construction pipeline in \cref{sec:benchmark_construction}. \cref{sec:benchmark_statistics} presents a statistical characterization of CSBench, highlighting its domain diversity, implementation complexity, and verification rigor.

\subsection{Task Definition}
\label{sec:task_definition}

Formally, a CSBench task instance is defined as a tuple $\mathcal{T} = (\mathcal{S}, \mathcal{C}_{init}, \mathcal{U}, \mathcal{E})$ with four components:
\begin{itemize}
    \item $\mathcal{S}$ (\textbf{Specification}): a Markdown file describing requirements and grading criteria;
    \item $\mathcal{C}_{init}$ (\textbf{Initial Context}): a scaffolded codebase (skeleton, headers, and build configuration);
    \item $\mathcal{U}$ (\textbf{Test Suite}): the canonical unit/integration tests serving as the grading oracle;
    \item $\mathcal{E}$ (\textbf{Environment}): an isolated docker container that hosts the full agent interaction context, including building, executing, and testing the code.
\end{itemize}

\paragraph{Model Input and Output.}
Given $(\mathcal{S}, \mathcal{C}_{init})$, a model produces a completed implementation $\mathcal{C}_{sol}$.
We standardize submissions by deriving a patch $\Delta = \text{diff}(\mathcal{C}_{init}, \mathcal{C}_{sol})$.

\paragraph{Execution-based Evaluation.}
We apply $\Delta$ to $\mathcal{C}_{init}$ and run $\mathcal{U}$ inside $\mathcal{E}$. We use the same environment $\mathcal{E}$ for agent execution and for final evaluation to ensure faithful, reproducible end-to-end assessment.
A task is considered solved if all tests in $\mathcal{U}$ pass after applying $\Delta$ within $\mathcal{E}$.

\subsection{Benchmark Construction}
\label{sec:benchmark_construction}

\paragraph{Overview.}
As illustrated in \cref{fig:construction_pipeline}, constructing CSBench involves a rigorous three-stage pipeline, \textbf{Sourcing}, \textbf{Annotation}, and \textbf{Quality Assurance}, designed to transform heterogeneous courseware into a standardized, execution-based benchmark.

Unlike \textit{passive} web scraping, our process employs an \textbf{adversarial, expert-in-the-loop workflow}.
Domain experts first act as {annotators} to instantiate the formal task tuple $\mathcal{T}=(\mathcal{S}, \mathcal{C}_{init}, \mathcal{U}, \mathcal{E})$, ensuring pedagogical fidelity.
These artifacts are then subjected to an \textbf{iterative refinement loop}, where independent human reviewers and automated agents challenge the tasks against strict executability and non-triviality criteria.
This multi-layered validation guarantees that every released task is not only technically reproducible but also robust against shortcut solutions.

\textbf{Phase 1: Sourcing.}
We first source candidate course projects from a diverse array of online open-source computer science courses from top-tier universities. We prioritize projects that provide a non-trivial, multi-file starter scaffold ($\mathcal{C}_{init}$) rather than single scripts.

\textbf{Phase 2: Annotation.}
For each sourced project, we perform a rigorous annotation process to produce a complete task instance in quadruple $(\mathcal{S}, \mathcal{C}_{\text{init}}, \mathcal{U}, \mathcal{E})$. The annotation process is carried out with two independent groups to enforce \textbf{adversarial and iterative} quality control:

\textit{I. Annotators} with domain expertise construct task artifacts by (i) refactoring original project files into a runnable scaffold $\mathcal{C}_{\text{init}}$ under a canonical layout; (ii) rewriting the assignment handout into a self-contained specification $\mathcal{S}$ while removing non-essential information; (iii) adapting official test suites when available or authoring new ones ($\mathcal{U}$); and (iv) configuring a Dockerized environment $\mathcal{E}$ for build-and-test execution. Finally, they need to submit a golden patch to verify test solvability.

\textit{II. Reviewers} audit annotators' outputs and provide actionable failure reports (e.g., missing requirements in $\mathcal{S}$, wrong layout in $\mathcal{C}_{\text{init}}$, insufficient coverage in $\mathcal{U}$, or failed builds in $\mathcal{E}$). Annotators revise the quadruple accordingly, iterating until it meets the benchmark construction criteria.

\textbf{Phase 3: Quality Assurance.}
Finally, each task undergoes a three-stage quality assurance process to secure executability, spec fidelity, and test trustworthiness:

\textit{I. Automated checks.} We verify canonical layout, Docker reproducibility, and task solvability: the test suite must fail on the initial scaffold $\mathcal{C}_{\text{init}}$ and pass 100\% after applying the golden patch.

\textit{II. AI-assisted auditing.}
For tasks passing automation, we employ LLM-based auditors to scan for subtle qualitative defects. The models analyze the $(\mathcal{S}, \mathcal{U})$ pair to flag potential information leakage, specification ambiguities, or insufficient test coverage that simple scripts cannot detect.

\textit{III. Human-in-the-loop decision.} Finally, experts serve as the ultimate decision gate, synthesizing outputs from the previous stages. We perform: (i) Audit Verification: manually reviewing AI-generated flags to confirm genuine spec inconsistencies or leakage risks; and (ii) Runtime Stability Inspection: examining execution logs from the golden run to identify and resolve environmental instabilities that automated boolean checks might overlook.

This three-stage construction process results in a pool of 237 verified tasks. 
From this pool, we carefully select 100 core tasks to form the final evaluation set of CSBench, balancing difficulty, category distribution, and annotation quality. 
This scale is comparable to recent agentic coding benchmarks~\citep{merrill2026terminal,zhou2026featurebench,thai2025swe}, while allowing us to maintain strict expert curation and execution-based quality control.

\subsection{Benchmark Characterization}
\label{sec:benchmark_statistics}
We characterize CSBench through a statistical analysis along four dimensions corresponding to its key design features: broad domain coverage, extensive specification context, substantial construction effort, and rigorous verification.

\textbf{Broad-scope domain coverage.} 
\cref{fig:category&language} illustrates CSBench's subject and programming-language coverage. CSBench spans four high-level topics (System, AI, Application and Others) in computer science education and 19 sub-topics, ensuring broad and systematic domain representation. The 100-task evaluation set is drawn from 43 courses and covers 9 programming languages: Python, C, C++, JavaScript, Java, Go, Lisp, Julia, and Rust. While Python (41\%) and C/C++ (39\%) form the core, reflecting the needs of AI-oriented and systems-oriented tasks, CSBench additionally includes other mainstream languages such as Java, JavaScript, and Go.

\begin{table*}[t]
    \centering
    \caption{Dataset statistics comparison: SWE-Bench~\citep{Jimenez2023SWEbenchCL}, Multi-SWE-Bench~\citep{zan2025multi}, and CSBench. Dashes indicate non-reported data. CSBench source counts exclude tests, data files, and build artifacts; detailed counting conventions are given in Appendix~\ref{app:detailed_statistics}.}
    {\small\setlength{\tabcolsep}{3pt}
    \begin{tabularx}{\textwidth}{ll *{6}{Y}}
    \toprule
    \multirow{2}{*}{\textbf{Metric}} & \multirow{2}{*}{\textbf{Attribute}} & \multicolumn{2}{c}{\textbf{SWE-Bench}} & \multicolumn{2}{c}{\textbf{Multi-SWE-Bench}} & \multicolumn{2}{c}{\textbf{CSBench}} \\
    \cmidrule(lr){3-4} \cmidrule(lr){5-6} \cmidrule(lr){7-8}
    & & Mean & Max & Mean & Max & Mean & Max \\
    \midrule
    Task Specification & Length & 195.1 & 4477 & 550.6 & -- & 676.2 & 5693 \\
    \midrule
    \multirow{2}{*}{Initial Code} 
    & \# Files (non-test) & 3,010 & 5,890 & 3817.5 & -- & 40.6 & 604 \\
    & \# Lines (non-test) & 438K & 886K & 178.6K & -- & 6.3K & 149.8K \\
    \midrule
    \multirow{3}{*}{Gold Patch}
    & \# Lines edited & 32.8 & 5888 & 163.3 & -- & 902.3 & 25K \\
    & \# Hunks edited & -- & -- & 11.7 & -- & 28.1 & 1530 \\
    & \# Files edited & 1.7 & 31 & 4.89 & -- & 10.6 & 363 \\
    \midrule
    \multirow{2}{*}{Test Suite}
    & \# Fail to Pass & 9.1 & 1633 & 6.26 & -- & 13.7 & 63 \\
    & \# Total & 120.8 & 9459 & 2241.2 & -- & 17.1 & 68 \\
    \bottomrule
    \end{tabularx}
    }
    \label{tab:benchmark-comparison}
\end{table*}

\textbf{Extensive specification context.}
CSBench tasks impose a heavy cognitive load on context understanding. 
As shown in Table~\ref{tab:benchmark-comparison}, starter repositories contain an average of 6.3K nonblank source lines and up to 149.8K, requiring agents to navigate existing code and integrate changes across files. 
Similarly, specifications average over 500 words with a maximum of nearly 5,700, demanding precise adherence to extensive, multi-file requirements.

\textbf{Substantial construction effort.}
Solutions in CSBench require structural architecture changes rather than localized patches. 
The average reference solution modifies 902.3 lines of code across 10.6 files. 
This represents an order-of-magnitude increase in implementation density compared to maintenance benchmarks like SWE-Bench (Table~\ref{tab:benchmark-comparison}), where a golden patch edits 32.8 lines across 1.7 files on average. This gap highlights CSBench's focus on large-scale \emph{system construction} rather than simple feature implementation.

\textbf{Rigorous execution-based verification.}
CSBench emphasizes Fail-to-Pass tests as the core verification signal. 
As shown in Table~\ref{tab:benchmark-comparison}, CSBench contains 13.7 Fail-to-Pass tests out of 17.1 total tests on average, a much higher proportion than maintenance benchmarks such as SWE-Bench and Multi-SWE-Bench. 
This design prevents agents from relying on existing passing logic and instead requires them to implement missing functionality that satisfies execution-based tests.

\section{Experiments}
\label{sec:experiments}
We evaluate current coding agents on CSBench for project-level system construction from specifications. We first describe the experimental setup, then present category-wise main results, followed by a Pass@k analysis of performance under additional sampling attempts.

\subsection{Settings}
\label{sec:experimental_settings}
We evaluate state-of-the-art models on CSBench using a minimalist agent loop equipped with stateless Bash execution and file editing tools. To ensure consistency across diverse models, we employ the default API parameters for each provider (e.g., temperature and top\_p) and utilize a standardized system prompt derived from the OpenHands agent to guide the problem-solving workflow. 

All experiments are conducted within isolated Docker containers to eliminate environmental discrepancies. Each task is allowed up to 50 interaction rounds, and we report the \textbf{average performance across 4 independent runs (avg@4)} for each instance. The agent is expected to autonomously handle environment feedback, such as compilation errors, through iterative reflection. A task is marked as failed if it exceeds the round limit and does not pass the unit test.

\begin{table}[t]
\caption{Main results. Numbers in parentheses indicate the number of tasks in each category. All results are reported as the average of 4 independent runs (avg@4). \\}
\centering
\small
\begin{tabular}{lccccc}
\toprule
\textbf{Model} & \multicolumn{4}{c}{\textbf{Category}} & \textbf{Overall} \\
\cmidrule(lr){2-5}
 & \topicSystem{System} & \topicAI{AI} & \topicApplication{Application} & \topicOthers{Others} &  \\
\midrule
Minimax-M2 & \score{6.52}\,\textcolor{gray}{\scriptsize $\pm$ 4.07} & \score{9.17}\,\textcolor{gray}{\scriptsize $\pm$ 4.33} & \score{11.84}\,\textcolor{gray}{\scriptsize $\pm$ 4.36} & \score{10.00}\,\textcolor{gray}{\scriptsize $\pm$ 10.00} & \score{8.50}\,\textcolor{gray}{\scriptsize $\pm$ 2.29} \\
Kimi-K2 & \score{5.98}\,\textcolor{gray}{\scriptsize $\pm$ 0.94} & \score{17.50}\,\textcolor{gray}{\scriptsize $\pm$ 1.44} & \score{9.21}\,\textcolor{gray}{\scriptsize $\pm$ 5.74} & \score{5.00}\,\textcolor{gray}{\scriptsize $\pm$ 8.66} & \score{10.00}\,\textcolor{gray}{\scriptsize $\pm$ 1.22} \\
Qwen3-Coder-Plus & \score{6.52}\,\textcolor{gray}{\scriptsize $\pm$ 3.44} & \score{16.67}\,\textcolor{gray}{\scriptsize $\pm$ 6.24} & \score{9.21}\,\textcolor{gray}{\scriptsize $\pm$ 5.74} & \score{10.00}\,\textcolor{gray}{\scriptsize $\pm$ 10.00} & \score{10.25}\,\textcolor{gray}{\scriptsize $\pm$ 3.56} \\
GLM-4.7 & \score{10.87}\,\textcolor{gray}{\scriptsize $\pm$ 3.44} & \score{20.83}\,\textcolor{gray}{\scriptsize $\pm$ 2.76} & \score{18.42}\,\textcolor{gray}{\scriptsize $\pm$ 4.56} & \score{15.00}\,\textcolor{gray}{\scriptsize $\pm$ 8.66} & \score{15.50}\,\textcolor{gray}{\scriptsize $\pm$ 1.66} \\
Seed-1.8 & \score{15.22}\,\textcolor{gray}{\scriptsize $\pm$ 3.07} & \score{11.67}\,\textcolor{gray}{\scriptsize $\pm$ 2.89} & \score{26.32}\,\textcolor{gray}{\scriptsize $\pm$ 7.44} & \score{15.00}\,\textcolor{gray}{\scriptsize $\pm$ 16.58} & \score{16.25}\,\textcolor{gray}{\scriptsize $\pm$ 2.68} \\
DeepSeek-V3.2 & \score{11.41}\,\textcolor{gray}{\scriptsize $\pm$ 0.94} & \score{29.17}\,\textcolor{gray}{\scriptsize $\pm$ 5.95} & \score{38.16}\,\textcolor{gray}{\scriptsize $\pm$ 7.78} & \score{15.00}\,\textcolor{gray}{\scriptsize $\pm$ 16.58} & \score{22.00}\,\textcolor{gray}{\scriptsize $\pm$ 1.58} \\
Gemini-3-Pro & \bestscore{38.59}\,\textcolor{gray}{\scriptsize $\pm$ 4.71} & \score{35.00}\,\textcolor{gray}{\scriptsize $\pm$ 6.01} & \score{46.05}\,\textcolor{gray}{\scriptsize $\pm$ 2.28} & \score{40.00}\,\textcolor{gray}{\scriptsize $\pm$ 14.14} & \score{39.00}\,\textcolor{gray}{\scriptsize $\pm$ 1.22} \\
Claude-Sonnet-4.5 & \score{33.70}\,\textcolor{gray}{\scriptsize $\pm$ 3.26} & \score{46.67}\,\textcolor{gray}{\scriptsize $\pm$ 5.27} & \score{60.53}\,\textcolor{gray}{\scriptsize $\pm$ 2.63} & \bestscore{90.00}\,\textcolor{gray}{\scriptsize $\pm$ 10.00} & \score{45.50}\,\textcolor{gray}{\scriptsize $\pm$ 0.87} \\
GPT-5.2 & \score{34.78}\,\textcolor{gray}{\scriptsize $\pm$ 3.44} & \bestscore{70.00}\,\textcolor{gray}{\scriptsize $\pm$ 2.36} & \bestscore{77.63}\,\textcolor{gray}{\scriptsize $\pm$ 7.78} & \score{75.00}\,\textcolor{gray}{\scriptsize $\pm$ 16.58} & \bestscore{55.50}\,\textcolor{gray}{\scriptsize $\pm$ 2.06} \\
\bottomrule
\end{tabular}
\label{tab:model_performance}
\end{table}

\subsection{Main Results}

We report the main results of CSBench on mainstream commercial and closed-source models across four broad categories: \textbf{System}, \textbf{AI}, \textbf{Application}, and \textbf{Others}. This high-level grouping allows us to examine how current agents perform under different forms of computational demands, ranging from low-level architectural constraints in system programming to model implementation and data-processing workflows in AI tasks.

Table~\ref{tab:model_performance} summarizes the category-wise performance. Overall, the results reveal substantial variation across task categories, indicating that CSBench captures heterogeneous demands beyond a single notion of coding ability. Among the four categories, \textbf{System} tasks are challenging for leading models: even top-performing models such as GPT-5.2 and Claude-Sonnet-4.5 achieve strong results in \textbf{Application} (77.63\% and 60.53\%) and \textbf{AI} (70.00\% and 46.67\%), but drop to only 34.78\% and 33.70\% on \textbf{System} tasks, respectively. This contrast suggests that current agents remain less capable of constructing low-level systems that require architectural consistency, dependency management, and execution-aware debugging. Moreover, the overall performance remains far from saturated, with the best model, GPT-5.2, achieving only 55.50\% overall, highlighting CSBench as a challenging and informative benchmark for measuring progress in project-level system construction.

The results further reveal \textbf{distinct domain-specific strengths across model families}. While GPT-5.2 achieves the best overall performance, other models exhibit competitive or superior performance in specific categories. For example, Gemini-3-Pro outperforms GPT-5.2 in System tasks (38.59\% vs. 34.78\%). These variations indicate that model capability is not uniform across computer science domains, highlighting the importance of fine-grained evaluation beyond aggregate scores.

\begin{figure}[tbp]
  \centering
  \includegraphics[width=.7\linewidth,trim=20bp 636bp 20bp 1bp,clip]{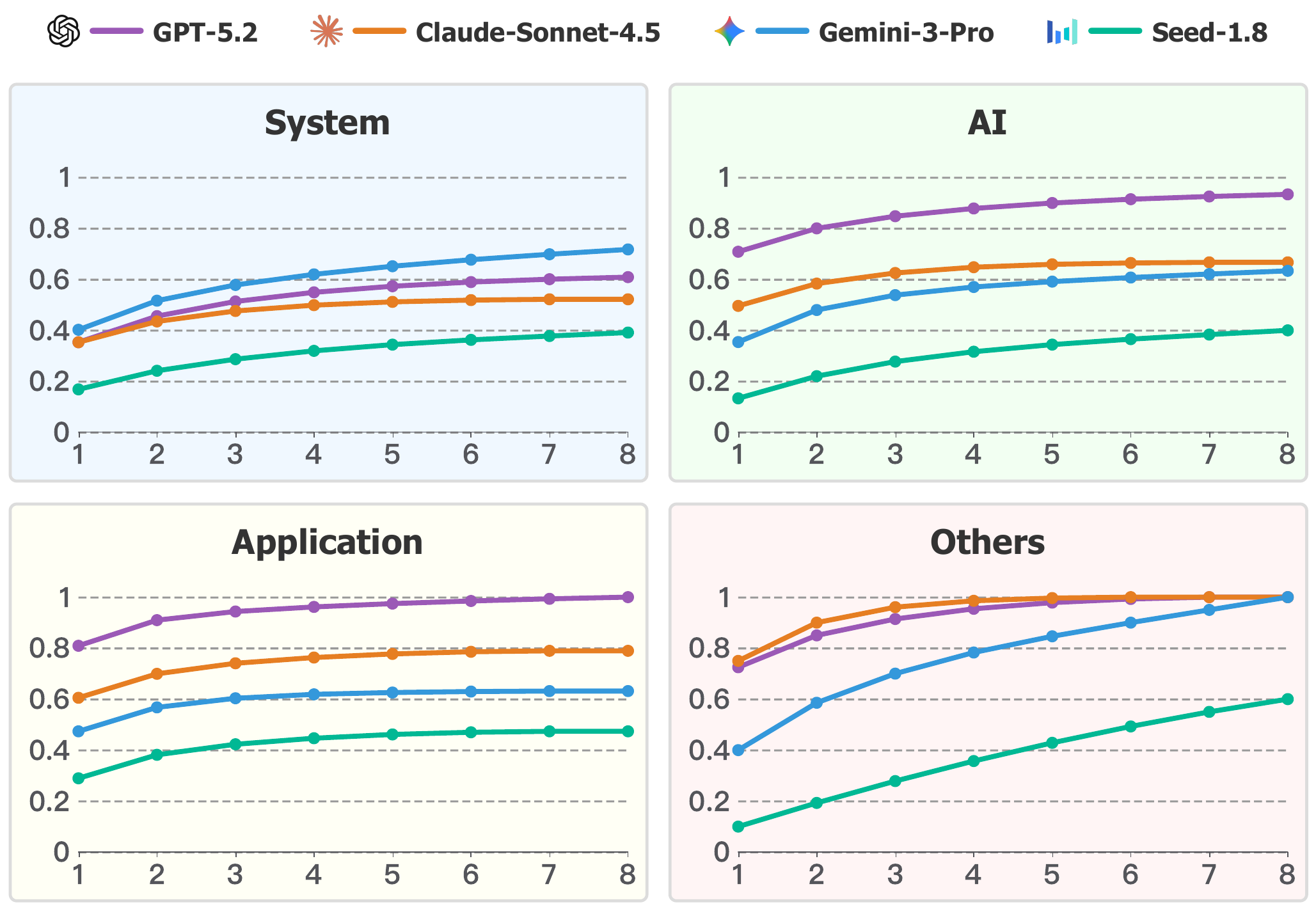}\\[2pt]
  \includegraphics[width=.247\linewidth,trim=3bp 322bp 497bp 58bp,clip]{figs/pass_at_k_2_2.pdf}\hfill
  \includegraphics[width=.247\linewidth,trim=498bp 322bp 3bp 58bp,clip]{figs/pass_at_k_2_2.pdf}\hfill
  \includegraphics[width=.247\linewidth,trim=3bp 7bp 497bp 373bp,clip]{figs/pass_at_k_2_2.pdf}\hfill
  \includegraphics[width=.247\linewidth,trim=498bp 7bp 3bp 373bp,clip]{figs/pass_at_k_2_2.pdf}
  \caption{Pass@k performance of representative models across different task categories.}
  \label{fig:pass_at_k}
\end{figure}

\subsection{Pass@k Analysis}
We report Pass@k metrics ($k=1$ to $8$) to evaluate whether models can solve CSBench tasks with multiple independent attempts. As shown in Figure~\ref{fig:pass_at_k}, performance generally improves with larger $k$, showing that additional sampling can recover some failures from individual trajectories. However, the gains are highly category-dependent: GPT-5.2 maintains a clear lead in \textbf{AI} and \textbf{Application}, whereas Gemini-3-Pro consistently performs best in \textbf{System} tasks across all $k$ values. System performance nevertheless remains far from saturated at the largest evaluated sampling budget ($k=8$), so repeated sampling does not close the observed gap within this tested range.

\section{Analysis}
To further investigate the model's performance on system-level tasks, we conduct a qualitative analysis of the problem-solving results. This analysis comprises three main components: (i) detailed performance analysis of each sub-category; (ii) an AI-annotated failure pattern analysis; and (iii) a case study of individual solution trajectories.

\begin{table*}[t]
\centering
\caption{Model Performance Scores across 19 fine-grained computer science topics in CSBench. Models are ordered by their overall performance. All results are reported as the average of 4 independent runs (avg@4). Topics with fewer than five tasks are diagnostic only.}
\label{tab:topic_performance}
\small
\renewcommand{\arraystretch}{1.6}
\resizebox{\textwidth}{!}{

\begin{tabular}{lccccccccc}
\toprule
\textbf{Topic} & \textbf{GPT-5.2} & \textbf{Claude-Sonnet-4.5} & \textbf{Gemini-3-Pro} & \textbf{DeepSeek-V3.2} & \textbf{Seed-1.8} & \textbf{GLM-4.7} & \textbf{Qwen3-Coder-Plus} & \textbf{Kimi-K2} & \textbf{Minimax-M2} \\
 & \small (OpenAI) & \small (Anthropic) & \small (Google) & \small (DeepSeek) & \small (ByteDance) & \small (Zhipu AI) & \small (Alibaba) & \small (Moonshot AI) & \small (MiniMax) \\
\midrule
\topicSystem{Operating System} & \score{37.50}\,\textcolor{gray}{\scriptsize $\pm$ 9.32} & \bestscore{45.83}\,\textcolor{gray}{\scriptsize $\pm$ 4.17} & \score{27.08}\,\textcolor{gray}{\scriptsize $\pm$ 9.08} & \score{8.33}\,\textcolor{gray}{\scriptsize $\pm$ 5.89} & \score{8.33}\,\textcolor{gray}{\scriptsize $\pm$ 8.33} & \score{10.42}\,\textcolor{gray}{\scriptsize $\pm$ 6.91} & \score{2.08}\,\textcolor{gray}{\scriptsize $\pm$ 3.61} & \score{6.25}\,\textcolor{gray}{\scriptsize $\pm$ 3.61} & \score{2.08}\,\textcolor{gray}{\scriptsize $\pm$ 3.61} \\
\topicSystem{System Fundamentals} & \score{19.44}\,\textcolor{gray}{\scriptsize $\pm$ 9.21} & \score{30.56}\,\textcolor{gray}{\scriptsize $\pm$ 9.21} & \bestscore{33.33}\,\textcolor{gray}{\scriptsize $\pm$ 11.11} & \score{16.67}\,\textcolor{gray}{\scriptsize $\pm$ 9.62} & \score{13.89}\,\textcolor{gray}{\scriptsize $\pm$ 4.81} & \score{2.78}\,\textcolor{gray}{\scriptsize $\pm$ 4.81} & \score{2.78}\,\textcolor{gray}{\scriptsize $\pm$ 4.81} & \score{0.00}\,\textcolor{gray}{\scriptsize $\pm$ 0.00} & \score{0.00}\,\textcolor{gray}{\scriptsize $\pm$ 0.00} \\
\topicSystem{System Security} & \score{50.00}\,\textcolor{gray}{\scriptsize $\pm$ 0.00} & \bestscore{54.17}\,\textcolor{gray}{\scriptsize $\pm$ 7.22} & \score{41.67}\,\textcolor{gray}{\scriptsize $\pm$ 14.43} & \score{4.17}\,\textcolor{gray}{\scriptsize $\pm$ 7.22} & \score{16.67}\,\textcolor{gray}{\scriptsize $\pm$ 20.41} & \score{16.67}\,\textcolor{gray}{\scriptsize $\pm$ 0.00} & \score{0.00}\,\textcolor{gray}{\scriptsize $\pm$ 0.00} & \score{8.33}\,\textcolor{gray}{\scriptsize $\pm$ 8.33} & \score{16.67}\,\textcolor{gray}{\scriptsize $\pm$ 0.00} \\
\topicSystem{Computer Network} & \bestscore{60.00}\,\textcolor{gray}{\scriptsize $\pm$ 0.00} & \score{35.00}\,\textcolor{gray}{\scriptsize $\pm$ 8.66} & \score{50.00}\,\textcolor{gray}{\scriptsize $\pm$ 10.00} & \score{20.00}\,\textcolor{gray}{\scriptsize $\pm$ 0.00} & \score{15.00}\,\textcolor{gray}{\scriptsize $\pm$ 8.66} & \score{15.00}\,\textcolor{gray}{\scriptsize $\pm$ 16.58} & \score{10.00}\,\textcolor{gray}{\scriptsize $\pm$ 10.00} & \score{15.00}\,\textcolor{gray}{\scriptsize $\pm$ 8.66} & \score{20.00}\,\textcolor{gray}{\scriptsize $\pm$ 14.14} \\
\topicSystem{Parallel \& Distributed System} & \score{20.00}\,\textcolor{gray}{\scriptsize $\pm$ 14.14} & \score{20.00}\,\textcolor{gray}{\scriptsize $\pm$ 0.00} & \bestscore{45.00}\,\textcolor{gray}{\scriptsize $\pm$ 8.66} & \score{20.00}\,\textcolor{gray}{\scriptsize $\pm$ 0.00} & \score{25.00}\,\textcolor{gray}{\scriptsize $\pm$ 8.66} & \score{15.00}\,\textcolor{gray}{\scriptsize $\pm$ 8.66} & \score{30.00}\,\textcolor{gray}{\scriptsize $\pm$ 17.32} & \score{15.00}\,\textcolor{gray}{\scriptsize $\pm$ 8.66} & \score{5.00}\,\textcolor{gray}{\scriptsize $\pm$ 8.66} \\
\topicSystem{Computer Architecture} & \score{31.25}\,\textcolor{gray}{\scriptsize $\pm$ 10.83} & \score{31.25}\,\textcolor{gray}{\scriptsize $\pm$ 20.73} & \bestscore{62.50}\,\textcolor{gray}{\scriptsize $\pm$ 12.50} & \score{12.50}\,\textcolor{gray}{\scriptsize $\pm$ 12.50} & \score{37.50}\,\textcolor{gray}{\scriptsize $\pm$ 12.50} & \score{25.00}\,\textcolor{gray}{\scriptsize $\pm$ 17.68} & \score{12.50}\,\textcolor{gray}{\scriptsize $\pm$ 12.50} & \score{0.00}\,\textcolor{gray}{\scriptsize $\pm$ 0.00} & \score{12.50}\,\textcolor{gray}{\scriptsize $\pm$ 21.65} \\
\topicSystem{Database System} & \score{31.25}\,\textcolor{gray}{\scriptsize $\pm$ 27.24} & \score{0.00}\,\textcolor{gray}{\scriptsize $\pm$ 0.00} & \bestscore{43.75}\,\textcolor{gray}{\scriptsize $\pm$ 10.83} & \score{0.00}\,\textcolor{gray}{\scriptsize $\pm$ 0.00} & \score{6.25}\,\textcolor{gray}{\scriptsize $\pm$ 10.83} & \score{0.00}\,\textcolor{gray}{\scriptsize $\pm$ 0.00} & \score{0.00}\,\textcolor{gray}{\scriptsize $\pm$ 0.00} & \score{0.00}\,\textcolor{gray}{\scriptsize $\pm$ 0.00} & \score{0.00}\,\textcolor{gray}{\scriptsize $\pm$ 0.00} \\
\topicSystem{Compiler} & \bestscore{25.00}\,\textcolor{gray}{\scriptsize $\pm$ 43.30} & \score{0.00}\,\textcolor{gray}{\scriptsize $\pm$ 0.00} & \score{0.00}\,\textcolor{gray}{\scriptsize $\pm$ 0.00} & \score{0.00}\,\textcolor{gray}{\scriptsize $\pm$ 0.00} & \score{0.00}\,\textcolor{gray}{\scriptsize $\pm$ 0.00} & \score{0.00}\,\textcolor{gray}{\scriptsize $\pm$ 0.00} & \score{0.00}\,\textcolor{gray}{\scriptsize $\pm$ 0.00} & \score{0.00}\,\textcolor{gray}{\scriptsize $\pm$ 0.00} & \score{0.00}\,\textcolor{gray}{\scriptsize $\pm$ 0.00} \\
\topicAI{Deep Learning} & \bestscore{71.15}\,\textcolor{gray}{\scriptsize $\pm$ 8.38} & \score{46.15}\,\textcolor{gray}{\scriptsize $\pm$ 5.44} & \score{36.54}\,\textcolor{gray}{\scriptsize $\pm$ 6.38} & \score{19.23}\,\textcolor{gray}{\scriptsize $\pm$ 8.60} & \score{13.46}\,\textcolor{gray}{\scriptsize $\pm$ 3.33} & \score{17.31}\,\textcolor{gray}{\scriptsize $\pm$ 3.33} & \score{13.46}\,\textcolor{gray}{\scriptsize $\pm$ 9.99} & \score{17.31}\,\textcolor{gray}{\scriptsize $\pm$ 6.38} & \score{0.00}\,\textcolor{gray}{\scriptsize $\pm$ 0.00} \\
\topicAI{Machine Learning} & \bestscore{94.44}\,\textcolor{gray}{\scriptsize $\pm$ 5.56} & \score{72.22}\,\textcolor{gray}{\scriptsize $\pm$ 5.56} & \score{36.11}\,\textcolor{gray}{\scriptsize $\pm$ 16.43} & \score{47.22}\,\textcolor{gray}{\scriptsize $\pm$ 12.11} & \score{11.11}\,\textcolor{gray}{\scriptsize $\pm$ 11.11} & \score{38.89}\,\textcolor{gray}{\scriptsize $\pm$ 5.56} & \score{22.22}\,\textcolor{gray}{\scriptsize $\pm$ 0.00} & \score{25.00}\,\textcolor{gray}{\scriptsize $\pm$ 9.21} & \score{30.56}\,\textcolor{gray}{\scriptsize $\pm$ 14.43} \\
\topicAI{Introductory AI} & \bestscore{50.00}\,\textcolor{gray}{\scriptsize $\pm$ 11.79} & \score{25.00}\,\textcolor{gray}{\scriptsize $\pm$ 8.33} & \score{41.67}\,\textcolor{gray}{\scriptsize $\pm$ 8.33} & \score{33.33}\,\textcolor{gray}{\scriptsize $\pm$ 11.79} & \score{12.50}\,\textcolor{gray}{\scriptsize $\pm$ 13.82} & \score{8.33}\,\textcolor{gray}{\scriptsize $\pm$ 8.33} & \score{20.83}\,\textcolor{gray}{\scriptsize $\pm$ 13.82} & \score{12.50}\,\textcolor{gray}{\scriptsize $\pm$ 7.22} & \score{0.00}\,\textcolor{gray}{\scriptsize $\pm$ 0.00} \\
\topicAI{Machine Learning System} & \bestscore{12.50}\,\textcolor{gray}{\scriptsize $\pm$ 21.65} & \score{0.00}\,\textcolor{gray}{\scriptsize $\pm$ 0.00} & \score{0.00}\,\textcolor{gray}{\scriptsize $\pm$ 0.00} & \score{0.00}\,\textcolor{gray}{\scriptsize $\pm$ 0.00} & \score{0.00}\,\textcolor{gray}{\scriptsize $\pm$ 0.00} & \score{0.00}\,\textcolor{gray}{\scriptsize $\pm$ 0.00} & \score{0.00}\,\textcolor{gray}{\scriptsize $\pm$ 0.00} & \score{0.00}\,\textcolor{gray}{\scriptsize $\pm$ 0.00} & \score{0.00}\,\textcolor{gray}{\scriptsize $\pm$ 0.00} \\
\topicApplication{Web Development} & \bestscore{85.71}\,\textcolor{gray}{\scriptsize $\pm$ 14.29} & \score{75.00}\,\textcolor{gray}{\scriptsize $\pm$ 6.19} & \score{25.00}\,\textcolor{gray}{\scriptsize $\pm$ 6.19} & \score{17.86}\,\textcolor{gray}{\scriptsize $\pm$ 11.85} & \score{7.14}\,\textcolor{gray}{\scriptsize $\pm$ 7.14} & \score{21.43}\,\textcolor{gray}{\scriptsize $\pm$ 7.14} & \score{3.57}\,\textcolor{gray}{\scriptsize $\pm$ 6.19} & \score{3.57}\,\textcolor{gray}{\scriptsize $\pm$ 6.19} & \score{14.29}\,\textcolor{gray}{\scriptsize $\pm$ 0.00} \\
\topicApplication{Data Structure \& Algorithm} & \bestscore{93.75}\,\textcolor{gray}{\scriptsize $\pm$ 10.83} & \score{81.25}\,\textcolor{gray}{\scriptsize $\pm$ 10.83} & \score{50.00}\,\textcolor{gray}{\scriptsize $\pm$ 17.68} & \score{62.50}\,\textcolor{gray}{\scriptsize $\pm$ 12.50} & \score{43.75}\,\textcolor{gray}{\scriptsize $\pm$ 10.83} & \score{25.00}\,\textcolor{gray}{\scriptsize $\pm$ 17.68} & \score{37.50}\,\textcolor{gray}{\scriptsize $\pm$ 21.65} & \score{25.00}\,\textcolor{gray}{\scriptsize $\pm$ 17.68} & \score{31.25}\,\textcolor{gray}{\scriptsize $\pm$ 20.73} \\
\topicApplication{Programming Fundamentals} & \score{50.00}\,\textcolor{gray}{\scriptsize $\pm$ 17.68} & \score{75.00}\,\textcolor{gray}{\scriptsize $\pm$ 17.68} & \bestscore{100.00}\,\textcolor{gray}{\scriptsize $\pm$ 0.00} & \score{81.25}\,\textcolor{gray}{\scriptsize $\pm$ 10.83} & \score{68.75}\,\textcolor{gray}{\scriptsize $\pm$ 20.73} & \score{25.00}\,\textcolor{gray}{\scriptsize $\pm$ 0.00} & \score{0.00}\,\textcolor{gray}{\scriptsize $\pm$ 0.00} & \score{12.50}\,\textcolor{gray}{\scriptsize $\pm$ 12.50} & \score{0.00}\,\textcolor{gray}{\scriptsize $\pm$ 0.00} \\
\topicApplication{Data Science} & \bestscore{66.67}\,\textcolor{gray}{\scriptsize $\pm$ 0.00} & \score{0.00}\,\textcolor{gray}{\scriptsize $\pm$ 0.00} & \score{33.33}\,\textcolor{gray}{\scriptsize $\pm$ 0.00} & \score{8.33}\,\textcolor{gray}{\scriptsize $\pm$ 14.43} & \score{0.00}\,\textcolor{gray}{\scriptsize $\pm$ 0.00} & \score{0.00}\,\textcolor{gray}{\scriptsize $\pm$ 0.00} & \score{0.00}\,\textcolor{gray}{\scriptsize $\pm$ 0.00} & \score{0.00}\,\textcolor{gray}{\scriptsize $\pm$ 0.00} & \score{0.00}\,\textcolor{gray}{\scriptsize $\pm$ 0.00} \\
\topicApplication{Software Engineering} & \bestscore{100.00}\,\textcolor{gray}{\scriptsize $\pm$ 0.00} & \score{0.00}\,\textcolor{gray}{\scriptsize $\pm$ 0.00} & \score{0.00}\,\textcolor{gray}{\scriptsize $\pm$ 0.00} & \score{0.00}\,\textcolor{gray}{\scriptsize $\pm$ 0.00} & \score{0.00}\,\textcolor{gray}{\scriptsize $\pm$ 0.00} & \score{0.00}\,\textcolor{gray}{\scriptsize $\pm$ 0.00} & \score{0.00}\,\textcolor{gray}{\scriptsize $\pm$ 0.00} & \score{0.00}\,\textcolor{gray}{\scriptsize $\pm$ 0.00} & \score{0.00}\,\textcolor{gray}{\scriptsize $\pm$ 0.00} \\
\topicOthers{Computer Graphics} & \score{58.33}\,\textcolor{gray}{\scriptsize $\pm$ 27.64} & \bestscore{100.00}\,\textcolor{gray}{\scriptsize $\pm$ 0.00} & \score{50.00}\,\textcolor{gray}{\scriptsize $\pm$ 16.67} & \score{25.00}\,\textcolor{gray}{\scriptsize $\pm$ 27.64} & \score{16.67}\,\textcolor{gray}{\scriptsize $\pm$ 16.67} & \score{25.00}\,\textcolor{gray}{\scriptsize $\pm$ 14.43} & \score{16.67}\,\textcolor{gray}{\scriptsize $\pm$ 16.67} & \score{8.33}\,\textcolor{gray}{\scriptsize $\pm$ 14.43} & \score{16.67}\,\textcolor{gray}{\scriptsize $\pm$ 16.67} \\
\topicOthers{Mathematics} & \bestscore{100.00}\,\textcolor{gray}{\scriptsize $\pm$ 0.00} & \score{75.00}\,\textcolor{gray}{\scriptsize $\pm$ 25.00} & \score{25.00}\,\textcolor{gray}{\scriptsize $\pm$ 25.00} & \score{0.00}\,\textcolor{gray}{\scriptsize $\pm$ 0.00} & \score{12.50}\,\textcolor{gray}{\scriptsize $\pm$ 21.65} & \score{0.00}\,\textcolor{gray}{\scriptsize $\pm$ 0.00} & \score{0.00}\,\textcolor{gray}{\scriptsize $\pm$ 0.00} & \score{0.00}\,\textcolor{gray}{\scriptsize $\pm$ 0.00} & \score{0.00}\,\textcolor{gray}{\scriptsize $\pm$ 0.00} \\
\midrule
\topic{Overall (100)} & \bestscore{55.50}\,\textcolor{gray}{\scriptsize $\pm$ 2.06} & \score{45.50}\,\textcolor{gray}{\scriptsize $\pm$ 0.87} & \score{39.00}\,\textcolor{gray}{\scriptsize $\pm$ 1.22} & \score{22.00}\,\textcolor{gray}{\scriptsize $\pm$ 1.58} & \score{16.25}\,\textcolor{gray}{\scriptsize $\pm$ 2.68} & \score{15.50}\,\textcolor{gray}{\scriptsize $\pm$ 1.66} & \score{10.25}\,\textcolor{gray}{\scriptsize $\pm$ 3.56} & \score{10.00}\,\textcolor{gray}{\scriptsize $\pm$ 1.22} & \score{8.50}\,\textcolor{gray}{\scriptsize $\pm$ 2.29} \\
\bottomrule
\end{tabular}

}
\end{table*}

\Needspace{5\baselineskip}
\subsection{Fine-Grained Performance Analysis}
Table~\ref{tab:topic_performance} presents performance across 19 fine-grained topics. Among topics with at least five tasks, GPT-5.2 reaches 94.44\% on Machine Learning ($N=9$), but only 37.50\% on Operating System ($N=12$), 20.00\% on Parallel \& Distributed System ($N=5$), and 19.44\% on System Fundamentals ($N=9$). Topics with $N<5$ are retained as diagnostic slices and are not used to support aggregate capability claims.

Difficulty also varies within the same broad category: GPT-5.2 performs relatively well on Computer Network (60.00\%, $N=5$) and System Security (50.00\%, $N=6$), while the larger or equal-sized system topics above remain substantially more challenging.

These disparities show that CSBench provides topic-level diagnostic signals beyond overall success rates, revealing which forms of computer science knowledge and implementation skills remain challenging for current coding agents.

\Needspace{14\baselineskip}
\subsection{Failure Pattern Analysis}
\begin{wrapfigure}[13]{r}{0.56\textwidth}
  \centering
  \includegraphics[width=0.95\linewidth]{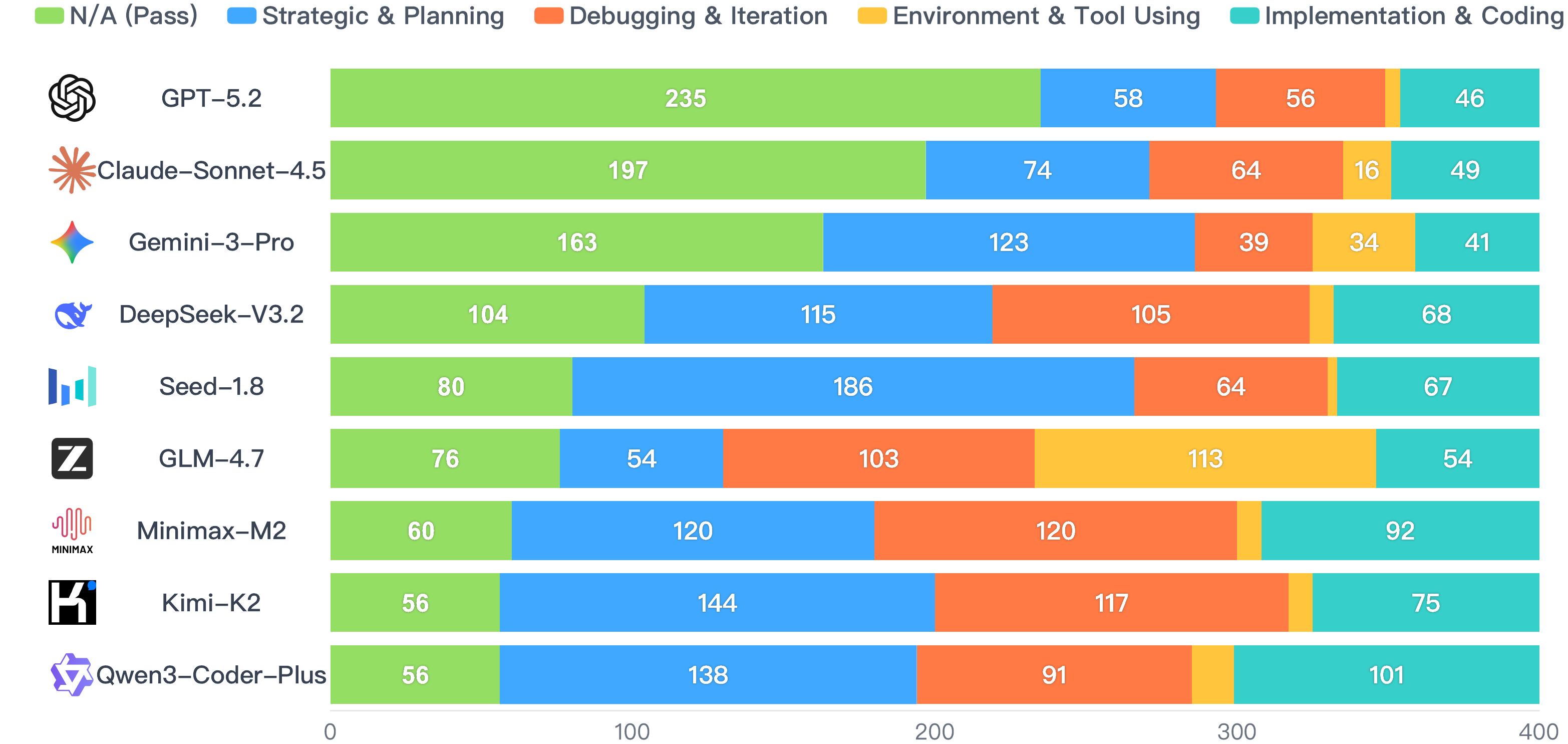}
  \caption{Distribution of primary failure categories.}
  \label{fig:failure_dist}
\end{wrapfigure}

To diagnose the root causes of performance failures in system construction, we annotate failed agent trajectories with a hierarchical failure taxonomy. As detailed in Section~\ref{sec:failure_pattern}, each failure is first assigned to one of four primary categories: 
(1) \textbf{Strategic \& Planning}, covering high-level failures in requirement understanding, design choice, or early termination; 
(2) \textbf{Implementation \& Coding}, covering concrete code-level errors such as compilation failures, runtime errors, or hallucinated APIs; 
(3) \textbf{Debugging \& Iteration}, covering failures to effectively use execution feedback, including repeated ineffective fixes or inability to locate the root cause;
(4) \textbf{Environment \& Tool Use}, covering failures in interacting with the execution environment or external tools. 
Each primary category is further decomposed into fine-grained sub-categories, which are detailed in \cref{tab:schema_mapping}.

As illustrated in Figure~\ref{fig:failure_dist}, \textbf{Strategic \& Planning} and \textbf{Debugging \& Iteration} constitute the majority of annotated failures in aggregate under the evaluated agent setting, accounting for 68.83\% of annotated failure instances. Within this setting, failures more often involve requirement tracking, planning, and iterative debugging than basic tool invocation. Appendix~\ref{app:failure_robustness} reports the corresponding domain- and topic-normalized robustness analysis.

Table~\ref{tab:top3_failure} further unpacks these setting-conditioned trends. Two patterns appear consistently across model families: \textbf{Stopped Too Early}, where agents terminate before satisfying all requirements, and \textbf{Failed to Locate Root Cause}, where agents observe failing tests or error logs but apply superficial fixes that do not address the underlying issue. \textbf{Missing Requirements} is also common, suggesting that agents often lose track of detailed specifications during long-horizon implementation. These distributions are descriptive of the evaluated agent setting rather than scaffold-independent model traits; Appendix~\ref{app:cross_scaffold} directly examines outcome sensitivity to a second scaffold.

\begin{table}[t]
\centering
\caption{The three most frequent patterns among 2,573 annotated failures.}
\label{tab:top3_failure}
\renewcommand{\arraystretch}{1.8}
\resizebox{\columnwidth}{!}{
\begin{tabular}{lrlll}
\toprule
\textbf{Model} & \textbf{\#Annotated failures} & \textbf{1st Frequent Failure} & \textbf{2nd Frequent Failure} & \textbf{3rd Frequent Failure} \\
\midrule
GPT-5.2 & 165 & \failDebug{Failed to Locate Root Cause (52)} & \failStrat{Stopped Too Early (37)} & \failImpl{Compilation Error (28)} \\
Claude-Sonnet-4.5 & 203 & \failDebug{Failed to Locate Root Cause (56)} & \failStrat{Missing Requirements (46)} & \failImpl{Compilation Error (25)} \\
Gemini-3-Pro & 237 & \failStrat{Stopped Too Early (75)} & \failStrat{Missing Requirements (38)} & \failDebug{Failed to Locate Root Cause (37)} \\
GLM-4.7 & 324 & \failEnv{Tool-call Format Error (112)} & \failDebug{Failed to Locate Root Cause (70)} & \failImpl{Runtime Error (34)} \\
Kimi-K2 & 344 & \failDebug{Failed to Locate Root Cause (107)} & \failStrat{Stopped Too Early (76)} & \failStrat{Missing Requirements (57)} \\
Minimax-M2 & 340 & \failDebug{Failed to Locate Root Cause (115)} & \failStrat{Missing Requirements (55)} & \failStrat{Stopped Too Early (53)} \\
DeepSeek-V3.2 & 296 & \failDebug{Failed to Locate Root Cause (91)} & \failStrat{Missing Requirements (59)} & \failStrat{Stopped Too Early (45)} \\
Seed-1.8 & 320 & \failStrat{Stopped Too Early (146)} & \failDebug{Failed to Locate Root Cause (56)} & \failImpl{Compilation Error (42)} \\
Qwen3-Coder-Plus & 344 & \failStrat{Stopped Too Early (85)} & \failDebug{Failed to Locate Root Cause (81)} & \failImpl{Compilation Error (62)} \\
\bottomrule
\end{tabular}
}
\end{table}

\Needspace{28\baselineskip}
\subsection{Case Study}
\begin{wrapfigure}[26]{r}{0.48\textwidth}
  \vspace{-\baselineskip}
  \centering
  \includegraphics[width=0.96\linewidth]{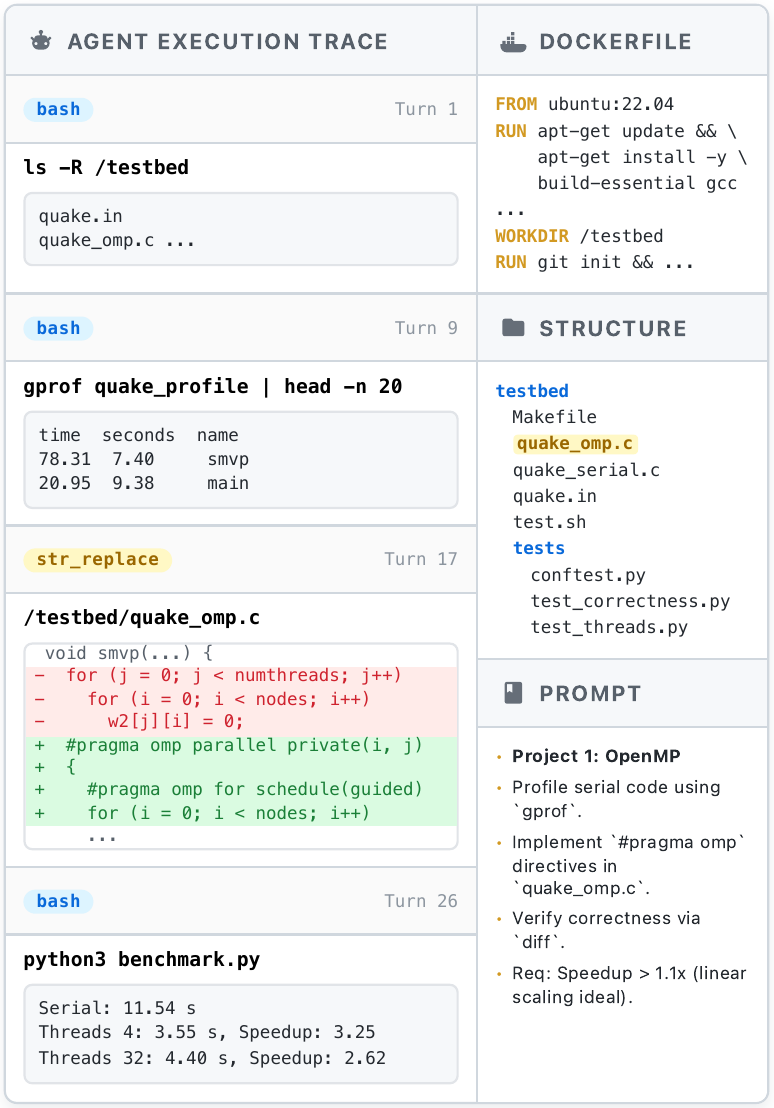}
  \caption{Gemini-3-Pro's trace on the OpenMP optimization task.}
  \label{fig:case_study}
\end{wrapfigure}
To demonstrate the system construction capabilities evaluated by CSBench, we analyze a representative trajectory of Gemini-3-Pro on a a high-performance computing task, which requires accelerating a serial simulation using OpenMP directives. As illustrated in Figure~\ref{fig:case_study}, the agent adopts a systematic engineering approach. It begins by exploring the repository structure to understand the build context and subsequently employs the \texttt{gprof} profiling tool to identify performance bottlenecks. The execution logs reveal that the agent correctly isolates the \texttt{smvp} function as the computational hotspot before attempting any modifications, focusing its optimization on the most performance-critical component.

Guided by the profiling results, the agent adds \texttt{\#pragma omp} directives to parallelize the loop in \texttt{smvp} within \texttt{quake\_omp.c}. It then runs \texttt{benchmark.py} and measures a $3.25\times$ speedup with four threads and a $2.62\times$ speedup with 32 threads. Both configurations exceed the task's $1.1\times$ target, although adding more threads reduces the speedup. This trajectory illustrates how CSBench evaluates an agent's ability to work autonomously across tasks, using profiling tools to diagnose a bottleneck, making a targeted code change, and verifying its effect.

\par\WFclear
\medskip
\FloatBarrier
\section{Related Work}

\textbf{Coding Agents.}
Coding agents have evolved from externally scaffolded workflows to models with increasingly internalized reasoning and self-improvement capabilities.
Early systems relied on \textit{agentic loops}, where LLMs are wrapped with planning, tool use, and repository interaction interfaces~\citep{Yang2024SWEagentAI, wang2024openhands, dong2024self}.
Representative frameworks such as SWE-agent~\citep{Yang2024SWEagentAI}, OpenHands~\citep{wang2024openhands}, and ChatDev~\citep{qian2024chatdev} established interactive environments~\citep{yang2023intercode, liu2023agentbench} for agents to navigate codebases, execute commands, and iteratively revise solutions.
However, such scaffolded workflows can suffer from context overflow, high latency, and brittle planning in long-horizon tasks.

Recent progress has shifted toward \textit{internalized reasoning}, where models learn to perform planning, search, and verification more directly through training.
Reasoning-oriented models such as DeepSeek-R1~\citep{guo2025deepseek}, OpenAI o1~\citep{jaech2024openai}, and Kimi K2~\citep{team2025kimi} use reinforcement learning and related post-training methods~\citep{da2025agent, tan2025scaling, karan2025reasoning, jin2025understanding} to reduce reliance on fragile external prompting.
Beyond fixed inference, recent work on \textit{self-evolving agents} explores systems that improve their own code, prompts, or policies through environment feedback and recursive reasoning~\citep{gao2025survey, zhang2025darwin, zhang2026evofsm, sampath2026adaptive, lin2025se}.
These trends motivate benchmarks that can evaluate whether agents acquire deeper, long-horizon engineering capabilities.

\textbf{Coding Agent Benchmarks.}
Code evaluation has shifted from isolated programming tasks~\citep{Chen2021EvaluatingLL,Austin2021ProgramSW,Hendrycks2021MeasuringCC,Li2022CompetitionlevelCG} to repository-scale agent benchmarks~\citep{Jimenez2023SWEbenchCL,Yang2024SWEagentAI,Xia2024AgentlessDL}.
SWE-bench~\citep{Jimenez2023SWEbenchCL} is a representative milestone, evaluating agents by resolving real-world GitHub issues in existing Python repositories.
Subsequent benchmarks extend this paradigm along multiple dimensions, including cross-file context retrieval~\citep{Liu2023RepoBenchBR,Ding2023CrossCodeEvalAD}, end-to-end repository generation~\citep{peng2026repogenesis,ding2025nl2repo}, feature development~\citep{wang2025swe}, multilingual issue resolution~\citep{zan2025multi}, and long-context planning~\citep{deng2025swe,rashid2025swe}.
Other specialized benchmarks target competitive programming~\citep{zheng2025livecodebench}, security-critical automation~\citep{lee2025sec, chen2025secureagentbench}, performance optimization~\citep{he2025swe, ouyang2025kernelbench}, and formal verification~\citep{yang2025verusage, xu2026agentic}.

RepoGenesis~\citep{peng2026repogenesis} and NL2Repo-Bench~\citep{ding2025nl2repo} study complementary empty-workspace construction regimes, whereas CSBench measures constrained implementation within executable starter repositories; Appendix~\ref{app:neighbor_comparison} provides a quantitative comparison.

Most existing repository-level benchmarks evaluate maintenance or adaptation within established codebases, such as bug fixing, issue resolution, or feature development.
CSBench instead focuses on project-level \textit{Spec-to-Code system construction}, where agents must build core computer science systems from rigorous specifications and scaffolded codebases.
By shifting the evaluation target from modifying existing software to constructing functional systems, CSBench directly probes agents' computer science mastery and their ability to translate specifications into implementations.

\section{Discussion}

\textbf{Intended Use.}
Researchers can use CSBench's domain-level results and failure taxonomy to identify weaknesses of coding agents in requirement tracking, implementation, and debugging, then evaluate targeted improvements to the agents under matched task sets, scaffolds, and interaction budgets. They can then compare success rates by domain and use the taxonomy to check whether the specific failure they aimed to address has become less frequent.

\textbf{Limitations.}
While CSBench provides a rigorous framework for system construction, it has certain limitations. First, our current evaluation primarily relies on binary pass/fail metrics based on unit tests. Future work could incorporate multi-dimensional rubrics such as code efficiency and architectural elegance. Second, tasks drawn from public courses may have appeared in model training data. Appendix~\ref{app:contamination_audit} reports our audit of textual overlap between task specifications and publicly available solutions. Lastly, our main results are conditioned on a shared scaffold, and the evaluation does not include a controlled human baseline. Future work could evaluate the same models across multiple scaffolds and compare agent performance with that of human experts on a common set of tasks.

\textbf{Broader Impact.}
The open release of CSBench supports reproducible evaluation and community contributions grounded in computer science education.
However, CSBench may raise concerns about reliance on AI-generated solutions in educational settings. We will therefore adopt responsible release practices and provide clear usage guidelines, while encouraging further research on governance mechanisms for AI-assisted learning.

\textbf{Conclusion.}
We introduce CSBench, a comprehensive benchmark that shifts the focus from software maintenance to project-level system construction. By curating a diverse and rigorous evaluation set from top-tier computer science curricula, we challenge LLM agents to bridge the gap between high-level specifications and low-level implementations. We evaluate 9 frontier LLMs on CSBench and find substantial variation across its construction domains. The failure analysis further shows that requirement-tracking and feedback-driven debugging failures are common in the evaluated trajectories. We hope CSBench will catalyze the development of next-generation coding agents capable of mastering the fundamental complexities of computer science.

\begin{ack}
This work is supported by the Natural Science Foundation of China (Grant No. 62376133) and sponsored by
Beijing Nova Program (20240484682).
\end{ack}

\bibliographystyle{iclr26}
\bibliography{main}

\appendix

{
\section{Supplementary Audits}

\subsection{Failure-distribution robustness}
\label{app:failure_robustness}

Table~\ref{tab:failure_robustness} retains the original set of 2,573 annotated failures and changes only the aggregation. Strategic \& Planning plus Debugging \& Iteration accounts for 68.83\% under micro averaging, 71.56\% after equal weighting of the four domains, and 67.28\% after equal weighting of the 19 topics. The qualitative conclusion is therefore not driven by the benchmark's unequal topic counts. Fine-grained topics with fewer than five tasks are used only diagnostically.

\begin{table}[htbp]
\centering
\caption{Primary failure-category distributions on the frozen annotation set. Domain rows are within-domain percentages; the last three rows use alternative aggregation rules.}
\label{tab:failure_robustness}
\small
\begin{tabular}{lrrrrr}
\toprule
Aggregation & $N$ & Planning & Implementation & Debugging & Environment \\
\midrule
System & 1,222 & 31.18 & 30.11 & 32.49 & 6.22 \\
AI & 768 & 45.83 & 19.14 & 24.74 & 10.29 \\
Application & 458 & 47.82 & 12.45 & 30.57 & 9.17 \\
Others & 125 & 48.00 & 16.80 & 25.60 & 9.60 \\
\midrule
Micro average & 2,573 & 39.33 & 23.05 & 29.50 & 8.12 \\
Domain-equalized & 4 domains & 43.21 & 19.63 & 28.35 & 8.82 \\
Topic-equalized & 19 topics & 37.70 & 24.15 & 29.57 & 8.57 \\
\bottomrule
\end{tabular}
\end{table}

\subsection{Failure-annotation protocol}

The 2,573 annotated failures exclude 137 non-pass reports omitted by the legacy collection filter. We labeled failed trajectories with \texttt{gemini-3-pro-preview-new} at temperature 0.1 in JSON mode. For each trajectory, the annotator received the task and attempt identifiers, at most the final 10,000 characters of the evaluation output, and the complete serialized agent conversation (which includes the task specification and tool feedback). The instruction requested one primary category, one fine-grained label, and a concise evidence-based reason identifying the root cause rather than the terminal symptom; in particular, turn-budget exhaustion itself was not an admissible cause. We additionally spot-checked a fixed, category-stratified sample of 50 annotations for obvious label/evidence mismatches. This was qualitative quality assurance rather than independent double annotation, so we do not claim inter-rater agreement. The sanitized verbatim prompt is included in the supplementary artifacts, and the complete taxonomy appears in Table~\ref{tab:schema_mapping}.

\subsection{Cross-scaffold sensitivity}
\label{app:cross_scaffold}

We compare the paper's unified OpenHands-derived loop with mini-SWE-agent v2.4.6 (commit \texttt{a83fcae}), an independently implemented bash-only loop. The balanced panel contains 50 tasks: 23 System, 15 AI, 9 Application, and 3 Others; 15 hard, 24 medium, and 11 easy; and all 19 topics across 9 languages. It retains the original locked 20-task pilot and was expanded without using results from the new scaffold. On the original scaffold, the reported panel preserves the full-benchmark order and closely matches its scores: GPT-5.2 52.5\% (full 55.5\%), DeepSeek-V3.2 23.0\% (22.0\%), and GLM-4.7 16.5\% (15.5\%). We run two independent attempts per task with the same specification, starter repository, Docker image, evaluator, model endpoint, temperature, output-token budget, and 50-call limit. We compare mini-SWE-agent avg@2 descriptively with the original avg@4; this is a scaffold-sensitivity check, not an estimate of a universal model score.

\begin{table}[htbp]
\centering
\caption{Cross-scaffold results on the fixed balanced 50-task panel. DeepSeek-V3.2 is an additional diagnostic run; all values are percentages.}
\label{tab:cross_scaffold}
\small
\begin{tabular}{lrrrr}
\toprule
Model & Full original & Panel original & mini-SWE-agent & Change \\
\midrule
GPT-5.2 & 55.5 & 52.5 & \textbf{60.0} & +7.5 \\
DeepSeek-V3.2 (diagnostic) & 22.0 & 23.0 & \textbf{20.0} & -3.0 \\
GLM-4.7 & 15.5 & 16.5 & \textbf{34.0} & +17.5 \\
\bottomrule
\end{tabular}
\end{table}

The scaffold effect is model-specific: GPT-5.2 and GLM-4.7 improve, whereas DeepSeek-V3.2 decreases by 3.0 percentage points. GPT-5.2 remains strongest, but GLM-4.7 and DeepSeek-V3.2 exchange order. We therefore condition absolute scores and lower-ranked comparisons on the scaffold rather than claiming exact rank invariance.

We also prepared a Harbor 0.20-compatible release using task schema 1.4, the existing CSBench images, a verifier, stable entry points, and an adapter for the original CodeAct loop. Standard Harbor tasks can subsequently be paired with built-in choices including mini-SWE-agent, SWE-agent, OpenHands/OpenHands SDK, Terminus, Aider, Goose, OpenCode, Codex, Gemini CLI, Qwen-Coder, Kimi, Cursor, Copilot, and Cline, or with custom Python/ACP agents. This interface enables broader future scaffold studies; it does not imply that every listed agent was evaluated here.

\subsection{Non-binary test diagnostics}
\label{app:per_test}

Strict full-suite success remains the primary metric because individual tests differ in scope and granularity. To expose partial progress, we additionally parse each original evaluator's final test summary. The task-normalized diagnostic first computes \texttt{passed/(passed+failed+errors)} per run and then averages parsed runs equally; the raw rate pools all parsed test outcomes and therefore gives more weight to larger suites. Evaluator runs without a completed test summary remain visible in the coverage column.

\begin{table}[htbp]
\centering
\caption{Non-binary diagnostics from the original 100-task, four-attempt evaluation.}
\label{tab:per_test}
\small
\begin{tabular}{lrrrr}
\toprule
Model & Binary avg@4 & Parsed summaries & Task-normalized & Raw per-test \\
\midrule
GPT-5.2 & 55.50 & 385/400 & 76.34 & 85.21 \\
Claude-Sonnet-4.5 & 45.50 & 381/400 & 76.26 & 81.80 \\
Gemini-3-Pro & 39.00 & 390/400 & 74.48 & 75.71 \\
DeepSeek-V3.2 & 22.00 & 380/400 & 61.27 & 66.24 \\
Seed-1.8 & 16.25 & 381/400 & 56.95 & 63.38 \\
Kimi-K2 & 10.00 & 380/400 & 49.80 & 59.01 \\
Minimax-M2 & 8.50 & 370/400 & 49.49 & 62.21 \\
GLM-4.7 & 15.50 & 382/400 & 48.36 & 54.36 \\
Qwen3-Coder-Plus & 10.25 & 381/400 & 46.83 & 58.75 \\
\bottomrule
\end{tabular}
\end{table}

The high-level separation is stable---GPT-5.2/Claude-Sonnet-4.5, then Gemini-3-Pro, DeepSeek-V3.2, and Seed-1.8---while the lower-score models reorder under partial-credit diagnostics. The larger absolute values confirm that binary scoring hides partial progress; the lower-tier reordering also shows why these values are complementary outcomes rather than calibrated fractions of system completion.

\subsection{Public-solution overlap audit}
\label{app:contamination_audit}

We audited all 20 tasks from the four widely replicated source families (MIT 6.S081, Full Stack Open, CS188 Pacman, and GAMES101). Before inspecting overlap results, we fixed two public repositories per family and resolved each to its latest commit before January 1, 2025. After excluding generated dependency lockfiles and vendored/build directories, the corpus contains 2,304 eligible code/text files and 2.03 million normalized tokens, yielding 19,088 overlapping 220-token windows (stride 110). For each task, same-family repositories form the aligned corpus and the other three families form negative controls.

As shown in Table~\ref{tab:overlap_audit}, verbatim overlap is small: mean maximum 5-gram containment is 0.98\%, its median is zero, and only 9 of 20 tasks have any 5-gram match. In a separate score sensitivity analysis, the 20 named tasks average 17.78\% across all nine models, versus 26.46\% for the remaining 80 tasks. Using all eight available non-named tasks in the same four topics as controls and weighting topic means by the named-set composition gives 41.53\%; the named set is lower for all nine models. No task or control was selected by score. These checks cannot determine membership in closed training corpora, but they find neither substantial specification copying nor anomalous performance inflation on the named sources.

\begin{table}[htbp]
\centering
\caption{Lexical overlap between 20 task specifications from the selected source families and a dated public-solution corpus. Values are normalized percentages.}
\label{tab:overlap_audit}
\small
\begin{tabular}{lrrrr}
\toprule
Metric & Aligned mean & Aligned median & Control mean & Difference \\
\midrule
5-gram containment & 0.98\% & 0.00\% & 0.00\% & +0.98 pp \\
Longest token-run ratio & 1.91\% & 0.00\% & 0.00\% & +1.91 pp \\
\bottomrule
\end{tabular}
\end{table}

\subsection{Comparison with neighboring construction benchmarks}
\label{app:neighbor_comparison}

Table~\ref{tab:neighbor_comparison} compares construction regimes by input, coverage, implementation scale, and evaluation protocol. The reported LOC values are directly useful as scale indicators, but occupy different roles: RepoGenesis reports an oracle implementation generated from scratch, NL2Repo-Bench reports the complete target repository, and CSBench reports both the provided code context and the required edit. We therefore present them side by side without collapsing them into a one-dimensional difficulty ranking.

\par\noindent\begin{minipage}{\linewidth}
\centering
\captionof{table}{Comparison with repository-construction benchmarks.}
\label{tab:neighbor_comparison}
\small
\begin{tabularx}{\textwidth}{lXXX}
\toprule
Benchmark & Input and construction regime & Evaluation scale and coverage & Oracle and metric \\
\midrule
RepoGenesis & API-oriented README plus dependency/build references; no source scaffold; generates a deployable microservice repository & 30 verified test repositories (106 including training); 18 application domains, 11 frameworks, Python/Java; oracle implementations span 100--2,426 LOC and 3--35 source files & 805 tests in the verified set; Pass@k, API Coverage, and Deployment Success \\
NL2Repo-Bench & Empty workspace; an approximately 18.8K-token textual blueprint supplies dependencies, the file tree, and API/signature descriptions; reconstructs a Python library & 104 repositories in 9 library categories; Python only; target repositories selected from 300--120K LOC; generated code can span 10K--50K tokens & Hidden upstream pytest; mean test-pass rate and fully-passed task count \\
CSBench & A 676.2-word course specification plus a non-trivial executable starter repository (6.3K mean nonblank source lines); completes a constrained core-CS system or component & 100 tasks from 43 courses across 19 topics and 9 programming languages; gold patches average 902.3 edited lines across 10.6 files & 17.1 tests/task (13.7 Fail-to-Pass); verified initial-fail/golden-pass invariant; strict task-level avg@4 \\
\bottomrule
\end{tabularx}
\end{minipage}\par
}

\FloatBarrier
\noindent\begin{minipage}{\linewidth}
\section{Failure Pattern Label}
\label{sec:failure_pattern}
Table~\ref{tab:schema_mapping} presents the hierarchical taxonomy used to annotate failed agent trajectories, comprising four primary categories and 13 fine-grained labels. Each annotated trajectory receives one primary category and one fine-grained label based on the failure evidence in its execution trace. Figure~\ref{fig:failure_dist} summarizes the distribution of primary categories, while Table~\ref{tab:top3_failure} reports the most frequent fine-grained failure patterns for each model.

\begingroup
\centering
\captionof{table}{Categorization Schema for Failure Pattern}
\label{tab:schema_mapping}
\resizebox{\columnwidth}{!}{
\begin{tabularx}{\textwidth}{>{\raggedright\arraybackslash}p{3.5cm} >{\raggedright\arraybackslash}p{4cm} X}
\toprule
\textbf{Primary Category} & \textbf{Sub-Category} & \textbf{Description / Explanation} \\
\midrule
\textbf{Strategic \& Planning} & Missing Requirements & Omission of specific constraints or functional requirements from the spec. \\
\cmidrule{2-3}
& Design Flaw & Architectural misalignment where the chosen approach conflicts with system structure. \\
\cmidrule{2-3}
& Stopped Too Early & Premature termination of the task before reaching a logical conclusion. \\
\midrule
\textbf{Implementation \& Coding} & Compilation Error & Failures occurring during the build, linking, or compilation phase. \\
\cmidrule{2-3}
& Runtime Error & Exceptions triggered during execution (e.g., Segfaults, Memory leaks). \\
\cmidrule{2-3}
& Usage of Non-existed Module/Function & Hallucinating APIs, libraries, or internal functions that do not exist. \\
\midrule
\textbf{Debugging \& Iteration} & Infinite Loop & Logic oscillation where the model repeats the same failed fix or code pattern. \\
\cmidrule{2-3}
& Side-Effect Fixing & Regressions where fixing one bug introduces new issues elsewhere. \\
\cmidrule{2-3}
& Failed to Locate Root Cause & Ineffective diagnosis leading to "surface-level" fixes that don't solve the core issue. \\
\cmidrule{2-3}
& Repetitive Submission & Stagnation characterized by submitting the same code multiple times without progress. \\
\midrule
\textbf{Environment \& Tool Using} & Env Corruption & Actions that destroy or misconfigure the underlying development environment. \\
\cmidrule{2-3}
& Test Hacking & Tampering with test suites or assertions to bypass validation rather than fixing the code. \\
\cmidrule{2-3}
& Tool Format Error & Failure to follow the specific output schema required by external tools or APIs. \\
\bottomrule
\end{tabularx}
}
\endgroup

\end{minipage}

\begin{figure*}[t]
  \centering
  \includegraphics[width=\linewidth]{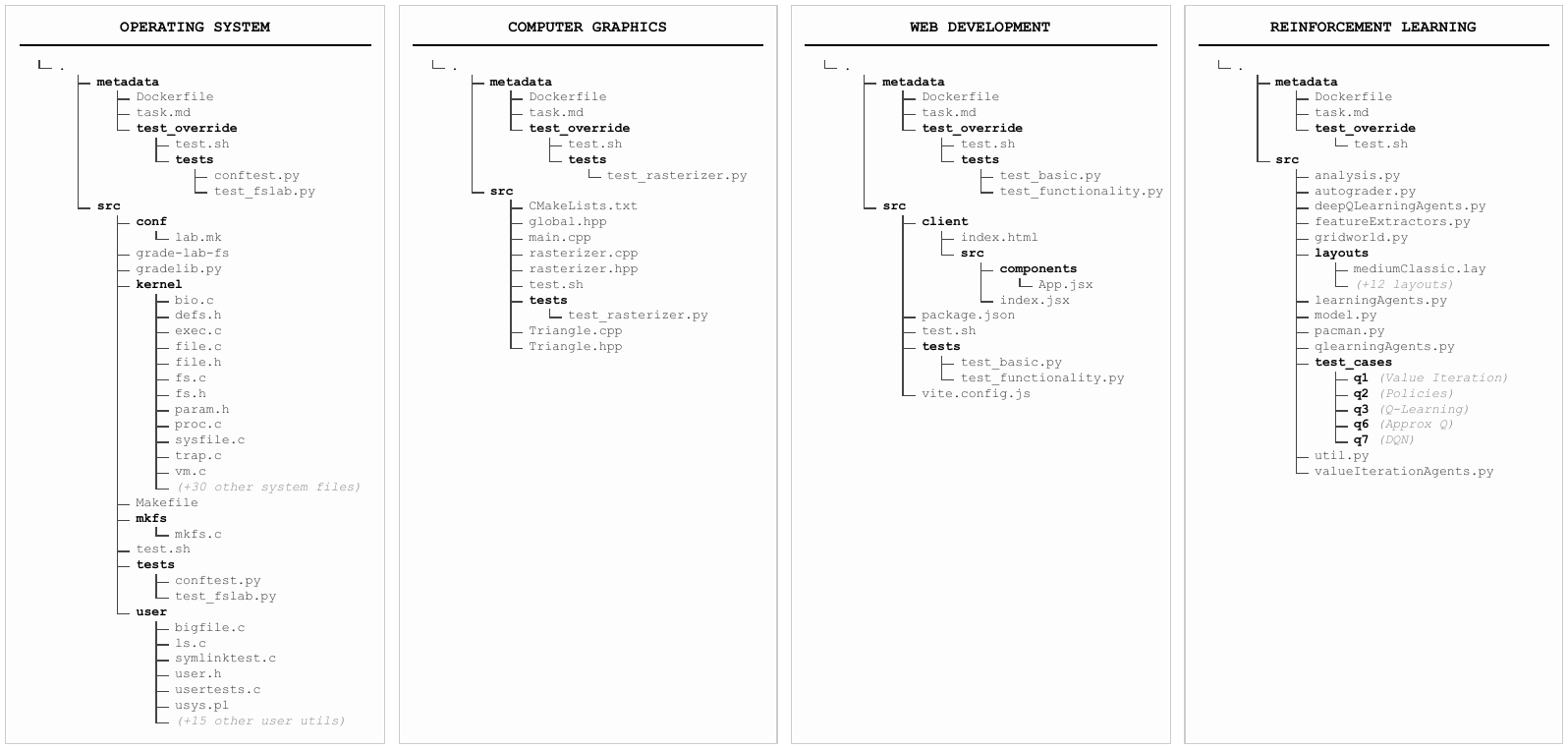}
  \caption{The standardized file structure of CSBench tasks across different domains.}
  \label{fig:file_structure}
\end{figure*}
\section{Extending Running Turns}

To assess sensitivity to the interaction budget, we extend the maximum agent turns from 50 to 100. As shown in Table~\ref{tab:topic_performance_100}, overall avg@4 increases by 11.0 percentage points for GPT-5.2, 18.0 for Sonnet-4.5, and 7.5 for Gemini-3-Pro under the same scaffold. Thus, additional interaction can materially improve task completion in this setting.

The topic-level effects are nevertheless heterogeneous. Most model--topic pairs improve or remain unchanged, while a few regress (e.g., Gemini-3-Pro decreases by 8.33 points on Operating Systems, and Sonnet-4.5 and Gemini-3-Pro each decrease by 10 points on Parallel \& Distributed Systems). With four attempts per task and several small topic subsets, we treat these changes as descriptive stochastic sensitivity rather than evidence of a universal ``scaling ceiling.'' The experiment shows that extra turns are not uniformly beneficial; it does not by itself identify the underlying cause or establish that additional interaction cannot compensate for domain-reasoning limitations.

\begin{table}[t]
\centering
\caption{Performance aggregated by topic after extending the turn limit to 100. All results are averages over four independent runs (avg@4). (*) marks subcategories with fewer than five tasks.}
\label{tab:topic_performance_100}
\small
\renewcommand{\arraystretch}{1.8}
\resizebox{\columnwidth}{!}{
\begin{tabular}{lcccccc}
\toprule
\textbf{Topic} & \multicolumn{2}{c}{\textbf{GPT-5.2}} & \multicolumn{2}{c}{\textbf{Sonnet-4.5}} & \multicolumn{2}{c}{\textbf{Gemini-3-Pro}} \\
\cmidrule(lr){2-3} \cmidrule(lr){4-5} \cmidrule(lr){6-7}
 & 50 turns & 100 turns & 50 turns & 100 turns & 50 turns & 100 turns \\
\midrule
\topicSystem{Operating System} & \score{37.50} & \score{43.75} \diffpos{6.25} & \score{45.83} & \bestscore{58.33} \diffpos{12.50} & \score{27.08} & \score{18.75} \diffneg{8.33} \\
\topicSystem{System Fundamentals} & \score{19.44} & \score{25.00} \diffpos{5.56} & \score{30.56} & \bestscore{52.78} \diffpos{22.22} & \score{33.33} & \score{44.44} \diffpos{11.11} \\
\topicSystem{System Security} & \score{50.00} & \bestscore{54.17} \diffpos{4.17} & \bestscore{54.17} & \bestscore{54.17} \diffzero & \score{41.67} & \score{37.50} \diffneg{4.17} \\
\topicSystem{Computer Network} & \score{60.00} & \bestscore{70.00} \diffpos{10.00} & \score{35.00} & \bestscore{70.00} \diffpos{35.00} & \score{50.00} & \score{60.00} \diffpos{10.00} \\
\topicSystem{Parallel \& Distributed System} & \score{20.00} & \score{20.00} \diffzero & \score{20.00} & \score{10.00} \diffneg{10.00} & \bestscore{45.00} & \score{35.00} \diffneg{10.00} \\
\topicSystem{Computer Architecture*} & \score{31.25} & \score{62.50} \diffpos{31.25} & \score{31.25} & \bestscore{81.25} \diffpos{50.00} & \score{62.50} & \score{68.75} \diffpos{6.25} \\
\topicSystem{Database System*} & \score{31.25} & \score{50.00} \diffpos{18.75} & \score{0.00} & \score{31.25} \diffpos{31.25} & \score{43.75} & \bestscore{56.25} \diffpos{12.50} \\
\topicSystem{Compiler*} & \score{25.00} & \bestscore{50.00} \diffpos{25.00} & \score{0.00} & \score{0.00} \diffzero & \score{0.00} & \score{0.00} \diffzero \\
\topicAI{Deep Learning} & \score{71.15} & \bestscore{92.31} \diffpos{21.15} & \score{46.15} & \score{76.92} \diffpos{30.77} & \score{36.54} & \score{40.38} \diffpos{3.85} \\
\topicAI{Machine Learning} & \score{94.44} & \bestscore{100.00} \diffpos{5.56} & \score{72.22} & \score{86.11} \diffpos{13.89} & \score{36.11} & \score{52.78} \diffpos{16.67} \\
\topicAI{Introductory AI} & \score{50.00} & \bestscore{62.50} \diffpos{12.50} & \score{25.00} & \score{50.00} \diffpos{25.00} & \score{41.67} & \score{41.67} \diffzero \\
\topicAI{Machine Learning System*} & \score{12.50} & \bestscore{25.00} \diffpos{12.50} & \score{0.00} & \score{12.50} \diffpos{12.50} & \score{0.00} & \score{0.00} \diffzero \\
\topicApplication{Web Development} & \score{85.71} & \bestscore{100.00} \diffpos{14.29} & \score{75.00} & \bestscore{100.00} \diffpos{25.00} & \score{25.00} & \score{96.43} \diffpos{71.43} \\
\topicApplication{Data Structure \& Algorithm*} & \bestscore{93.75} & \bestscore{93.75} \diffzero & \score{81.25} & \bestscore{93.75} \diffpos{12.50} & \score{50.00} & \score{50.00} \diffzero \\
\topicApplication{Programming Fundamentals*} & \score{50.00} & \score{87.50} \diffpos{37.50} & \score{75.00} & \score{93.75} \diffpos{18.75} & \bestscore{100.00} & \score{93.75} \diffneg{6.25} \\
\topicApplication{Data Science*} & \bestscore{66.67} & \bestscore{66.67} \diffzero & \score{0.00} & \score{0.00} \diffzero & \score{33.33} & \score{25.00} \diffneg{8.33} \\
\topicApplication{Software Engineering*} & \bestscore{100.00} & \bestscore{100.00} \diffzero & \score{0.00} & \score{50.00} \diffpos{50.00} & \score{0.00} & \score{0.00} \diffzero \\
\topicOthers{Computer Graphics*} & \score{58.33} & \score{66.67} \diffpos{8.33} & \bestscore{100.00} & \bestscore{100.00} \diffzero & \score{50.00} & \score{50.00} \diffzero \\
\topicOthers{Mathematics*} & \bestscore{100.00} & \score{87.50} \diffneg{12.50} & \score{75.00} & \score{50.00} \diffneg{25.00} & \score{25.00} & \score{50.00} \diffpos{25.00} \\
\midrule
\topic{Overall (100)} & \score{55.50} & \bestscore{66.50} \diffpos{11.00} & \score{45.50} & \score{63.50} \diffpos{18.00} & \score{39.00} & \score{46.50} \diffpos{7.50} \\
\bottomrule
\end{tabular}
}
\end{table}

\section{Detailed Task Examples}
To demonstrate the diversity of CSBench, we detail four representative tasks covering Operating Systems, Graphics, Web Development, and Artificial Intelligence. The standardized file structure across all tasks is illustrated in \cref{fig:file_structure}.

\paragraph{Operating System (xv6 File System)}\leavevmode\\
\textbf{Source:} Adapted from MIT 6.S081 (xv6 Labs). \\
\textbf{Task:} The agent must modify the xv6 kernel to overcome file system limitations. Requirements include implementing doubly-indirect blocks to support large files (up to 65,803 blocks) and adding a \texttt{symlink} system call with recursive path resolution and cycle detection.

\paragraph{Computer Graphics (Rasterization)}\leavevmode\\
\textbf{Source:} Adapted from GAMES101 Assignment 2. \\
\textbf{Task:} The goal is to implement a hardware rasterizer simulation in C++. Key requirements involve coding the \texttt{rasterize\_triangle} function, implementing Z-buffering via barycentric interpolation for depth testing, and optionally supporting Super-Sampling Anti-Aliasing (SSAA).

\paragraph{Web Development (React App)}\leavevmode\\
\textbf{Source:} Adapted from Full Stack Open (University of Helsinki). \\
\textbf{Task:} The agent is tasked with building a React application. Requirements include rendering data collections using \texttt{map()}, handling user input via controlled forms, and implementing asynchronous CRUD operations to interact with a remote backend server.

\paragraph{Reinforcement Learning (Pacman)}\leavevmode\\
\textbf{Source:} Adapted from UC Berkeley CS188 Project 6. \\
\textbf{Task:} The task involves implementing reinforcement learning algorithms for Gridworld and Pacman agents. The agent must code Value Iteration, Q-Learning, Approximate Q-Learning, and a Deep Q-Network (DQN) with gradient updates.

\FloatBarrier
\section{Detailed Statistics of CSBench}
\label{app:detailed_statistics}
We provide fine-grained statistics of CSBench stratified by \emph{task type} and \emph{programming language} in \cref{tab:stats-by-type} and \cref{tab:stats-by-language}. Specification length is measured in whitespace-separated words. Source size counts nonblank lines, including comments, in source files, headers, grammar files, build scripts, and notebook code cells; vendored source is included, while tests, data assets, binary files, and build artifacts are excluded. Source-file counts use the same inclusion rules. F2P denotes Fail-to-Pass tests. Total-test counts additionally include outcomes involving skipped tests. Overall means are computed over the same 100 tasks, with half-up rounding.

\paragraph{Cross-benchmark comparability.}
Table~\ref{tab:benchmark-comparison} reports SWE-bench and Multi-SWE-Bench statistics from their original publications alongside CSBench statistics computed using the source-file counting rules above. The figures characterize dataset scale; counting conventions may differ across benchmarks. Specification lengths are reported in words for SWE-bench and CSBench and in tokens for Multi-SWE-Bench.

\begin{table}[!htbp]
\centering
\small
\caption{CSBench Statistics by Task Type}
\label{tab:stats-by-type}
\setlength{\tabcolsep}{4pt}
\begin{tabular}{l|r|rr|rr|rr|r}
\toprule
\textbf{Type} & \textbf{Count} & \multicolumn{2}{c|}{\textbf{Spec Words}} & \multicolumn{2}{c|}{\textbf{Source LoC}} & \multicolumn{2}{c|}{\textbf{Patch Lines}} & \textbf{F2P Tests} \\
 & & Mean & Max & Mean & Max & Mean & Max & Mean \\
\midrule
System & 46 & 774.3 & 3828 & 11696.3 & 149792 & 1501.4 & 25247 & 11.8 \\
AI & 30 & 813.4 & 5693 & 1496.7 & 6906 & 484.1 & 1590 & 18.9 \\
Application & 19 & 296.8 & 1061 & 2542.6 & 14092 & 319.5 & 786 & 11.2 \\
Others & 5 & 393.4 & 824 & 458.0 & 896 & 113.4 & 245 & 10.8 \\
\midrule
\textbf{Total} & \textbf{100} & 676.2 & 5693 & 6335.3 & 149792 & 902.3 & 25247 & 13.7 \\
\bottomrule
\end{tabular}
\end{table}

\begin{table}[!htbp]
\centering
\small
\caption{CSBench Statistics by Programming Language}
\label{tab:stats-by-language}
\setlength{\tabcolsep}{4pt}
\begin{tabular}{l|r|rr|rr|rr|r}
\toprule
\textbf{Language} & \textbf{Count} & \multicolumn{2}{c|}{\textbf{Spec Words}} & \multicolumn{2}{c|}{\textbf{Source LoC}} & \multicolumn{2}{c|}{\textbf{Patch Lines}} & \textbf{F2P Tests} \\
 & & Mean & Max & Mean & Max & Mean & Max & Mean \\
\midrule
Python & 41 & 730.0 & 5693 & 1243.5 & 6906 & 417.1 & 1590 & 19.2 \\
C++ & 21 & 512.3 & 1664 & 17750.6 & 149792 & 1344.5 & 21244 & 8.6 \\
C & 18 & 747.2 & 3828 & 7201.3 & 15753 & 2094.4 & 25247 & 7.8 \\
JavaScript & 8 & 356.1 & 1061 & 1921.6 & 14092 & 311.9 & 490 & 18.0 \\
Java & 5 & 861.0 & 1858 & 10884.8 & 25180 & 499.2 & 775 & 2.6 \\
Go & 3 & 619.7 & 803 & 2321.0 & 2964 & 488.7 & 796 & 15.3 \\
Lisp & 2 & 277.0 & 330 & 105.5 & 122 & 181.0 & 292 & 9.0 \\
Julia & 1 & 824.0 & 824 & 234.0 & 234 & 119.0 & 119 & 41.0 \\
Rust & 1 & 3094.0 & 3094 & 2955.0 & 2955 & 251.0 & 251 & 2.0 \\
\midrule
\textbf{Total} & \textbf{100} & 676.2 & 5693 & 6335.3 & 149792 & 902.3 & 25247 & 13.7 \\
\bottomrule
\end{tabular}
\end{table}

\FloatBarrier
\section{Cost-Performance Analysis}
In addition to pass rates, inference cost is an important practical factor when deploying coding agents. 
Table~\ref{tab:cost_performance} reports the overall performance of each model together with its API pricing at the time of evaluation.
While stronger closed-source models generally achieve higher success rates, their inference costs vary substantially. 
For example, DeepSeek-V3.2 achieves a moderate overall score of 22.00\% with substantially lower input and output costs than most frontier models, whereas GPT-5.2 obtains the best overall performance but at a higher output-token cost.

\begin{table}[h]
\centering
\caption{
Cost-performance comparison of evaluated models on CSBench. 
Prices denote API costs per million tokens at the time of evaluation.
}
\label{tab:cost_performance}
\centering
\small
\begin{tabular}{lccc}
\toprule
\textbf{Model} & \textbf{Overall} & \textbf{Input Cost (\$/M tokens)} & \textbf{Output Cost (\$/M tokens)} \\
\midrule
Minimax-M2          & 8.50\%  & 0.10 & 0.10 \\
Kimi-K2             & 10.00\% & 0.55 & 2.19 \\
Qwen3-Coder-Plus    & 10.25\% & 0.54 & 2.16 \\
GLM-4.7             & 15.50\% & 0.41 & 1.92 \\
DeepSeek-V3.2       & 22.00\% & 0.30 & 0.45 \\
Gemini-3-Pro        & 39.00\% & 2.00 & 12.00 \\
Claude-Sonnet-4.5   & 45.50\% & 3.00 & 15.00 \\
GPT-5.2             & 55.50\% & 1.75 & 14.00 \\
\bottomrule
\end{tabular}
\end{table}

\section{Computational Resources}
\label{app:compute_resources}
Running a single full evaluation pass over CSBench typically costs between \$10 and \$200, depending on the model pricing and generated token usage.
On a machine with 128 CPU cores and sufficient memory, one pass over the dataset takes approximately 2 hours, excluding the time required to build Docker images.

\newpage
\input{checklist.tex}

\end{document}

%% file: checklist.tex
\section*{NeurIPS Paper Checklist}

\begin{enumerate}

\item {\bf Claims}
    \item[] Question: Do the main claims made in the abstract and introduction accurately reflect the paper's contributions and scope?
    \item[] Answer: \answerYes{} %
    \item[] Justification: The abstract and introduction accurately describe CSBench's scope, construction process, and main empirical findings without making unsupported claims beyond the presented experiments.

\item {\bf Limitations}
    \item[] Question: Does the paper discuss the limitations of the work performed by the authors?
    \item[] Answer: \answerYes{} %
    \item[] Justification: The paper already included a limitation section in the main text.

\item {\bf Theory assumptions and proofs}
    \item[] Question: For each theoretical result, does the paper provide the full set of assumptions and a complete (and correct) proof?
    \item[] Answer: \answerNA{} %
    \item[] Justification: The paper does not include theoretical results.

    \item {\bf Experimental result reproducibility}
    \item[] Question: Does the paper fully disclose all the information needed to reproduce the main experimental results of the paper to the extent that it affects the main claims and/or conclusions of the paper (regardless of whether the code and data are provided or not)?
    \item[] Answer: \answerYes{} %
    \item[] Justification: We include a brief description of experimental setup in Section~\ref{sec:experimental_settings}, which is sufficient to reproduce the main results of our experiments.

\item {\bf Open access to data and code}
    \item[] Question: Does the paper provide open access to the data and code, with sufficient instructions to faithfully reproduce the main experimental results, as described in supplemental material?
    \item[] Answer: \answerYes{} %
    \item[] Justification: We open-source our data and code in supplemental materials.

\item {\bf Experimental setting/details}
    \item[] Question: Does the paper specify all the training and test details (e.g., data splits, hyperparameters, how they were chosen, type of optimizer) necessary to understand the results?
    \item[] Answer: \answerYes{} %
    \item[] Justification: We clarify the details of our dataset and the testing scheme in Section 2 and 3.

\item {\bf Experiment statistical significance}
    \item[] Question: Does the paper report error bars suitably and correctly defined or other appropriate information about the statistical significance of the experiments?
    \item[] Answer: \answerYes{} %
    \item[] Justification: We report the standard deviations for our main experiments.

\item {\bf Experiments compute resources}
    \item[] Question: For each experiment, does the paper provide sufficient information on the computer resources (type of compute workers, memory, time of execution) needed to reproduce the experiments?
    \item[] Answer: \answerYes{} %
    \item[] Justification: We provide a discussion of compute resources in Appendix~\ref{app:compute_resources}.
    
\item {\bf Code of ethics}
    \item[] Question: Does the research conducted in the paper conform, in every respect, with the NeurIPS Code of Ethics \url{https://neurips.cc/public/EthicsGuidelines}?
    \item[] Answer: \answerYes{} %
    \item[] Justification: We have reviewed the Code of Ethics and made sure that our paper conforms to it.

\item {\bf Broader impacts}
    \item[] Question: Does the paper discuss both potential positive societal impacts and negative societal impacts of the work performed?
    \item[] Answer: \answerYes{} %
    \item[] Justification: We have discussed both positive and negative societal impacts in Section 6 of our paper.
    
\item {\bf Safeguards}
    \item[] Question: Does the paper describe safeguards that have been put in place for responsible release of data or models that have a high risk for misuse (e.g., pre-trained language models, image generators, or scraped datasets)?
    \item[] Answer: \answerNA{} %
    \item[] Justification: The paper poses no such risks.

\item {\bf Licenses for existing assets}
    \item[] Question: Are the creators or original owners of assets (e.g., code, data, models), used in the paper, properly credited and are the license and terms of use explicitly mentioned and properly respected?
    \item[] Answer: \answerYes{} %
    \item[] Justification: We declare the licenses of our released code and data, properly credit the original asset creators, and ensure that all reused materials comply with their respective licenses and terms of use.

\item {\bf New assets}
    \item[] Question: Are new assets introduced in the paper well documented and is the documentation provided alongside the assets?
    \item[] Answer: \answerYes{} %
    \item[] Justification: The new dataset introduced is well documented in our paper. Documentation is provided with the open-source assets.

\item {\bf Crowdsourcing and research with human subjects}
    \item[] Question: For crowdsourcing experiments and research with human subjects, does the paper include the full text of instructions given to participants and screenshots, if applicable, as well as details about compensation (if any)? 
    \item[] Answer: \answerNA{} %
    \item[] Justification: The paper does not involve crowdsourcing nor research with human subjects.

\item {\bf Institutional review board (IRB) approvals or equivalent for research with human subjects}
    \item[] Question: Does the paper describe potential risks incurred by study participants, whether such risks were disclosed to the subjects, and whether Institutional Review Board (IRB) approvals (or an equivalent approval/review based on the requirements of your country or institution) were obtained?
    \item[] Answer: \answerNA{} %
    \item[] Justification: The paper does not involve crowdsourcing nor research with human subjects.

\item {\bf Declaration of LLM usage}
    \item[] Question: Does the paper describe the usage of LLMs if it is an important, original, or non-standard component of the core methods in this research? Note that if the LLM is used only for writing, editing, or formatting purposes and does \emph{not} impact the core methodology, scientific rigor, or originality of the research, declaration is not required.
    \item[] Answer: \answerYes{} %
    \item[] Justification: Apart from writing and editing purposes, LLM is used for data auditing and is described in Section 2 of our paper. This does not affect the rigor and originality of our research.

\end{enumerate}